\documentclass[final,3p,times]{elsarticle}

\usepackage{epsfig}

\usepackage{amssymb}
\usepackage{amsthm}

\usepackage{amsmath,amsfonts}
\usepackage{subcaption}
\usepackage{color}
\usepackage{multirow} 

\usepackage{wrapfig}

\journal{Neurocomputing}

\begin{document}

\begin{frontmatter}



\title{Active Self-Semi-Supervised Learning for Few Labeled Samples}

\author[label1]{Ziting Wen \corref{cor1}}\ead{zwen4889@uni.sydney.edu.au}

\author[label1,label2]{Oscar Pizarro}\ead{oscar.pizarro@ntnu.no}
\affiliation[label1]{organization={Australian Centre for Robotics, The University of
Sydney},
            city={Sydney},
            country={Australia}}

\affiliation[label2]{organization={Department of Marine Technology, Norwegian University of Science and Technology},
            city={Trondheim},
            country={Norway}}

\author[label1]{Stefan Williams}\ead{stefan.williams@sydney.edu.au}

\cortext[cor1]{Corresponding author}

\begin{abstract}
Training deep models with limited annotations poses a significant challenge when applied to diverse practical domains. Employing semi-supervised learning alongside the self-supervised model offers the potential to enhance label efficiency. However, this approach faces a bottleneck in reducing the need for labels. We observed that the semi-supervised model disrupts valuable information from self-supervised learning when only limited labels are available. To address this issue, this paper proposes a simple yet effective framework, active self-semi-supervised learning (AS3L). AS3L bootstraps semi-supervised models with prior pseudo-labels (PPL). These PPLs are obtained by label propagation over self-supervised features. Based on the observations the accuracy of PPL is not only affected by the quality of features but also by the selection of the labeled samples. We develop active learning and label propagation strategies to obtain accurate PPL. Consequently, our framework can significantly improve the performance of models in the case of limited annotations while demonstrating fast convergence. On the image classification tasks across four datasets, our method outperforms the baseline by an average of 5.4\%. Additionally, it achieves the same accuracy as the baseline method in about 1/3 of the training time.

\end{abstract}




\end{frontmatter}


\section{Introduction}
\label{}

Part of the success of deep learning models arises from large amounts of labeled data~\cite{deng2009imagenet}. However, the high cost associated with acquiring large numbers of labels hinders the widespread application of deep learning models. This challenge is particularly pronounced in domains that demand expert annotations, such as medical images~\cite{rahmati2024redundant}, or biology images~\cite{alonso2019coralseg}. In response, researchers have dedicated considerable effort to exploring semi-supervised learning. Recent works have shown that these techniques can achieve similar accuracy to supervised learning with fewer annotations~\cite{sohn2020fixmatch,zhang2021flexmatch}.

However, existing semi-supervised learning techniques face bottlenecks in reducing labeled samples~\cite{kim2022propagation}. This is due to the common practice of employing model predictions as pseudo-labels for unlabeled samples, which are not always accurate. Training with these noisy pseudo-labels boosts the model's confidence in incorrect predictions, inducing the model to resist the correct information, and consequently compromising the performance of semi-supervised training. This issue is commonly referred to as confirmation bias~\cite{arazo2020pseudo}. To unlock the full potential of semi-supervised learning, many current techniques still rely on a substantial number of annotations to rectify these incorrect pseudo-labels. For instance, prevalent benchmarks in semi-supervised learning typically opt for 25, 100, and 400 annotations per class~\cite{berthelot2019mixmatch,xie2020unsupervised,rizve2020defense}, with some recent studies gradually reducing to 4 annotations per class~\cite{zhang2021flexmatch,li2021comatch}. As the number of annotations decreases, the semi-supervised learning struggles to correct erroneous pseudo-labels, resulting in a significant decline in its performance~\cite{kim2022propagation}.


\begin{figure}[!htbp]
  \centering
  
  \begin{subfigure}{0.35\textwidth}
    \includegraphics[width=\linewidth]{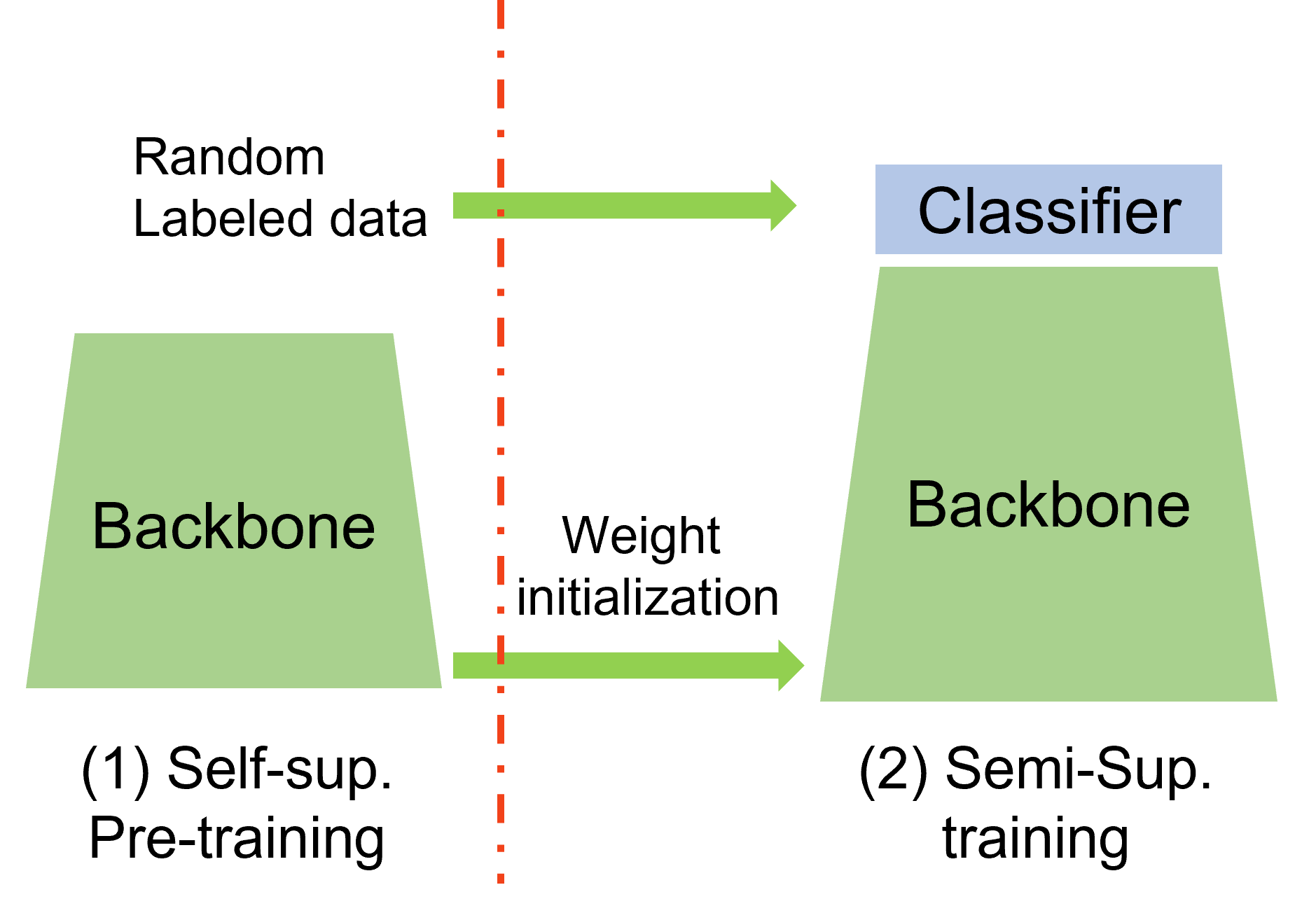}
    \caption{SelfMatch}
    \label{fig:intro_selfm}
  \end{subfigure}
  \begin{subfigure}{0.55\textwidth}
    \includegraphics[width=\linewidth]{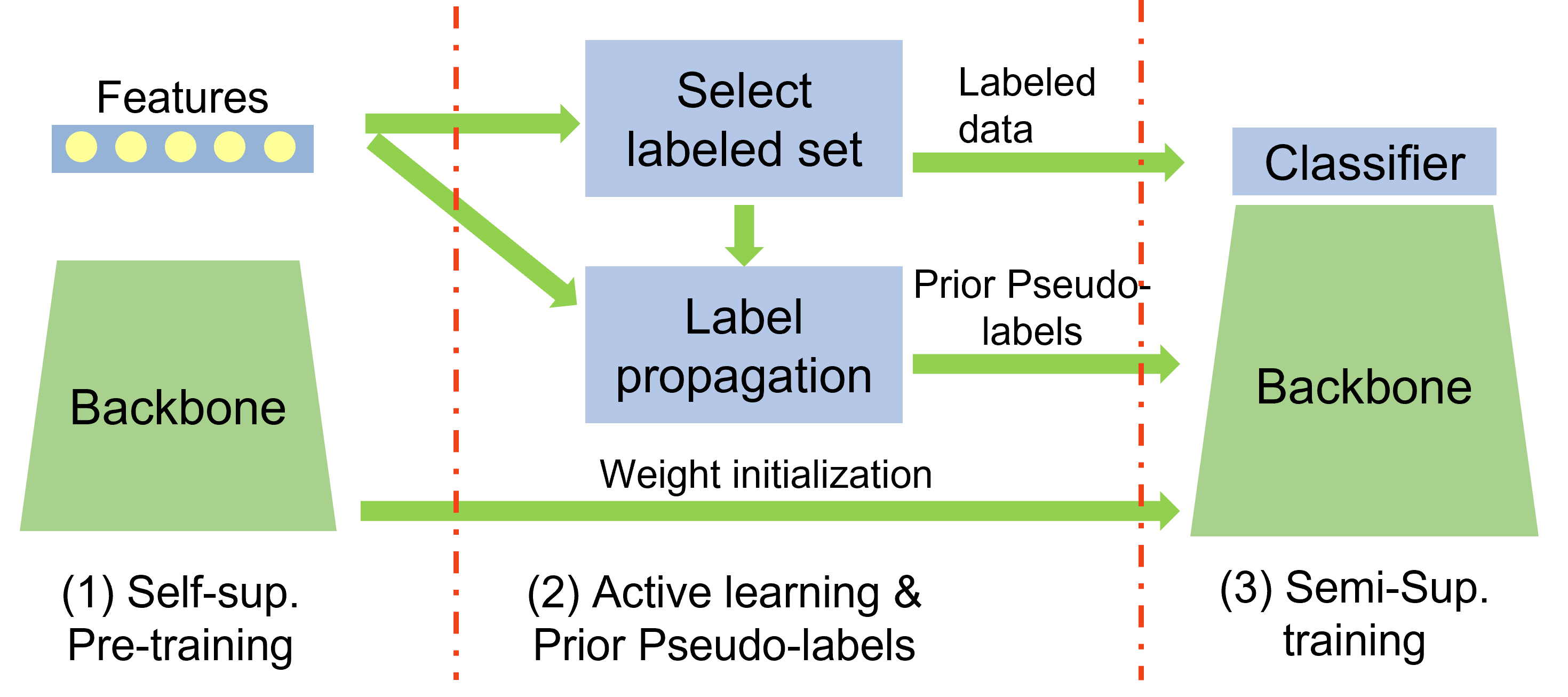}
    \caption{AS3L (ours)}
    \label{fig:intro_as3l}
  \end{subfigure}

  \caption{Pipeline of AS3L (Ours) and Existing Self-Semi-Supervised Learning Approache (SelfMatch)~\cite{kim2021selfmatch}. (a) SelfMatch involves self-supervised pre-training followed by semi-supervised fine-tuning, relying on weight initialization to benefit semi-supervised learning from self-supervised pre-training. (b) Beyond weight initialization, AS3L (ours) improves semi-supervised learning by selecting labeled samples and generating prior pseudo-labels based on self-supervised features, providing a better starting point for subsequent semi-supervised training.}
  \label{fig:intro_frame}
\end{figure}

Conducting semi-supervised training on top of self-supervised models holds great promise in improving this issue~\cite{xu2024revisiting}. Self-supervised training yields valuable representations for downstream tasks without any labels~\cite{caron2020unsupervised,cai2021exponential}. Initializing with a self-supervised model allows the semi-supervised model to swiftly generate accurate pseudo-labels and reduce the influence of erroneous pseudo-labels~\cite{kim2021selfmatch}, thereby enhancing final performance. However, with limited labeled samples, the semi-supervised model may not rapidly generate accurate pseudo-labels. These inaccurate pseudo-labels disrupt valuable information obtained from self-supervised learning (Sec.~\ref{sec:insight}), consequently diminishing the performance gains achieved through the initialization with self-supervised learning. In some cases, this initialization might not yield any performance improvements.

Motivated by this observation, we propose a more explicit method for transferring valuable information from self-supervised models to semi-supervised models using \textbf{Prior Pseudo-Labels (PPL)}. Our proposed framework, Active Self-Semi-Supervised Learning (AS3L) depicted in Fig.~\ref{fig:intro_frame}, aims to leverage self-supervised pre-training to enhance model performance. Specifically, we use PPL to guide semi-supervised learning. PPL is obtained through label propagation on self-supervised representations (Sec.~\ref{sec:prior_pslabel}). Additionally, we implement a switching mechanism to integrate PPL and model predictions early in training and gradually phase out PPL as training progresses (Sec.~\ref{sec:semisup_training}). This approach provides a strong starting point for semi-supervised training and prevents PPL from hindering the continuous updating of the pseudo-labels during the semi-supervised training process. Furthermore, considering that the accuracy of PPL depends on both feature quality and labeled sample selection, particularly when annotations are limited, we propose an active learning strategy. This strategy aims to improve the accuracy of PPL by optimizing the selection of labeled samples (Sec. \ref{sec:ss_active}).

Validated across four image classification datasets, our AS3L outperforms the state-of-the-art in most cases, particularly in scenarios with limited labels, and showcases accelerated convergence (Sec. \ref{sec:res}). The contributions of our work are summarized as follows: 1) In scenarios with scarce labeled data, our observation reveals that semi-supervised learning cannot effectively harness valuable information obtained from self-supervised training by weight initialization. 2) Motivated by the observation, we propose the AS3L framework, a novel approach to bootstrap semi-supervised training from a good initial point consisting of accurate PPL, actively selected labeled samples, and weights initialization from self-supervised training. This method readily integrates with existing semi-supervised learning approaches to extend the success of semi-supervised learning to scenarios with even fewer labeled data. Our proposed method outperforms SOTA algorithms in most cases with limited annotations. 3) We develop an active learning strategy that is tightly coupled to the AS3L framework, which can greatly improve the accuracy of PPL when there are few labeled samples, thereby helping AS3L work well with limited labels.

\section{Related Work}

\subsection{Semi-Supervised Learning}
\subsubsection{Semi-Supervised Learning from Scratch}

Semi-supervised training commonly leverages unlabeled samples through pseudo-labeling techniques~\cite{lee2013pseudo} and consistency regularization methods~\cite{tarvainen2017mean}. Pseudo-labeling involves using model predictions as targets for training unlabeled data, typically using only high-confidence predictions to reduce the impact of incorrect targets on semi-supervised training~\cite{rizve2020defense}. Consistency regularization aims to make the model's predictions consistent for perturbed inputs by the consistency loss. Various studies have explored different perturbation methods. For example, UDA uses basic image transformations such as flipping and cropping~\cite{xie2020unsupervised}, while VAT generates perturbed inputs through adversarial training~\cite{miyato2018virtual}. Recent studies propose combining pseudo-labeling and consistency regularization to enhance semi-supervised training performance~\cite{berthelot2019mixmatch}. For instance, \textbf{FixMatch} employs strong and weak image augmentations on the same input images, using high-confidence model predictions from the weakly augmented version as targets for the strongly augmented images~\cite{sohn2020fixmatch}. However, early in the training process, the model’s prediction confidence is low for many samples, leading to slow convergence in FixMatch. To address this issue, \textbf{FlexMatch} introduces dynamic confidence thresholds, gradually increasing them during training to balance convergence speed and pseudo-label accuracy~\cite{zhang2021flexmatch}.

However, in scenarios with limited annotated data, existing methods exhibit sub-optimal performance due to a lack of supervisory signals in the early stages of training and the impact of confirmation bias. Also, the issue of limited labels often leads to training instability, which often results in the optimal checkpoint being distant from the final training iteration. \textbf{Sel} introduces a metric designed to select high-performing checkpoints, to mitigate the challenge posed by the instability in semi-supervised training~\cite{kim2022propagation}.

\subsubsection{Semi-Supervised Learning with Self-Supervision} Through the training of proxy tasks, \textbf{self-supervised learning} leverages all available data to generate a valuable representation and model initialization weights for downstream tasks before acquiring annotations. These self-supervision signals provide additional information to semi-supervised training, improving its performance, especially when annotations are limited. Self-supervised tasks mainly include generative-based and contrastive-based methods~\cite{liu2021self}. Generative self-supervised learning trains an encoder-decoder model to reconstruct inputs from corrupted inputs. For instance, MAE trains a model to reconstruct input images with large randomly masked-out portions~\cite{he2022masked}, while DMAE reconstructs inputs from noisy, randomly masked images~\cite{wudenoising}. Contrastive training trains an encoder by comparing the similarities between the output features. Early contrastive training methods utilized the relationships between the overall input sample and its parts. For example, solving jigsaw involves randomly sampling patches from an input image and training the model to predict their relative positions~\cite{noroozi2016unsupervised}. Rotation prediction trains the model to predict the angle of randomly rotated inputs~\cite{DBLP:conf/iclr/GidarisSK18}. More recently, methods comparing features of different samples or various augmented versions of the same sample have been widely explored. SimCLR constructs positive pairs for contrastive learning by applying different augmentations to the same input, and negative pairs using a large batch of different samples, significantly improving self-supervised learning performance~\cite{chen2020simple}. However, this method requires many negative samples, leading to a computational bottleneck. To address this, recent methods explore only positive pairs for contrastive learning. \textbf{BYOL} uses a moving-average teacher model and an online model to perform contrastive learning~\cite{grill2020bootstrap}, while \textbf{SimSiam} directly uses contrastive loss to encourage similarity between features of the same input under different augmentations~\cite{chen2021exploring}.

How current semi-supervised training algorithms utilize self-supervised learning techniques can be divided into three categories. The first and most straightforward way is to propagate labels on self-supervised features directly~\cite{bai2021self}. However, self-supervised features are not perfect for a specific task so simple label propagation based on this feature does not produce ideal results. 

The second category of methods uses both semi-supervised loss and self-supervised loss to train the model. S4L trains a model with self-supervised loss (predict rotation~\cite{DBLP:conf/iclr/GidarisSK18}) and semi-supervised loss, then re-trains the model with the model's prediction~\cite{zhai2019s4l}. \textbf{CoMatch} adds contrastive self-supervised loss to the standard semi-supervised loss~\cite{li2021comatch}. \textbf{LESS} incorporates self-supervised signals into the semi-supervised training process through online clustering~\cite{lucas2022barely}. These methods improve the performance of semi-supervised training when dealing with extremely limited labeled samples. However, when the number of labels is limited, the poor pseudo-labels may affect the representation generated by the model, thereby hindering the effective utilization of self-supervision loss. Moreover, these methods still fail to address the challenge of effectively transferring knowledge from existing pre-trained models to semi-supervised models.

The third category of methods transfers information obtained from self-supervised training primarily relying on weight initialization. \textbf{SelfMatch} proposes initializing the model with self-supervised pre-training weights, followed by fine-tuning with existing semi-supervised training methods~\cite{kim2021selfmatch}. EMAN uses exponential moving averages of model and batch normalization parameters as a teacher model~\cite{cai2021exponential}. This approach slows the alteration of self-supervised pre-training weights, improving semi-supervised training performance. While self-supervised training provides commendable initial weights for the backbone, it does not give initial weights for classifiers due to the different loss and network architecture. In scenarios with limited labeled samples, poor pseudo-labels at the beginning of semi-supervised training can rapidly disrupt the good initial weights established through self-supervised learning. Therefore, this paper introduces a framework designed to facilitate the effective transfer of knowledge acquired during the self-supervised stage. This framework utilizes pseudo-labels generated by self-supervised features to guide the process of semi-supervised learning. Additionally, our approach can be combined with techniques like EMAN to yield enhanced results.

\subsection{Active Learning}

Most semi-supervised learning techniques select labeled samples by random stratified sampling~\cite{zhang2021flexmatch,li2021comatch}. As the number of labels decreases, the gap between labeled and unlabeled sample distributions increases, which worsens the performance of semi-supervised learning. It is straightforward to fuse active learning techniques to fill in the gap. \textbf{Active learning} assumes that different samples have different extents of influence on model accuracy. When we can only afford to label a fraction of samples, choosing samples with higher values to label can result in a more accurate model. One popular active learning approach selects samples for labeling based on the uncertainty of model predictions, with uncertainty measured using metrics like entropy~\cite{lewis1994heterogeneous}, BALD~\cite{gal2017deep} or learning indicator \cite{yoo2019learning}. Another approach aims to select representative samples for labeling. For example, Coreset selects samples that are least similar to the features of already labeled samples~\cite{sener2018active}. However, this method can easily identify outliers. To address this, some studies have explored selecting samples from high-density regions to avoid the influence of outliers~\cite{wu2019cost,wang2022cost,settles2009active}.

\subsubsection{Active Strategy for Semi-Supervised Learning}
Some researchers have tried to directly combine existing semi-supervised learning with these active strategies developed in the context of supervised learning, but the results were unexpectedly poor, even worse than random selection baseline~\cite{mittal2019parting}. To bridge this gap, some new active strategies have been designed in the background of semi-supervised learning. These strategies are more tightly coupled with existing semi-supervised learning approaches. Consistency-based methods~\cite{gao2020consistency} argue that we should choose samples that are hard for semi-supervised models, i.e. samples with inconsistent predictions for data augmentation. Guo et al. proposed to combine adversarial training and graph-based label propagation to select samples close to cluster boundaries with high uncertainty~\cite{guo2021semi}. However, these strategies are multiple-shot strategies, which means an unbearably high computational burden in semi-supervised learning scenarios. The substantial training cost generated by multi-round active semi-supervised learning contradicts the overarching goal of minimizing the cost required to train a powerful model throughout the entire research endeavor.

\subsubsection{Single-shot Active strategy}
Single-shot active learning strategies request all labels at a single time, which may help us benefit from active learning with little additional computational burden. Until now, little attention has been paid to this field. Several novel active learning approaches have focused on selecting labeled samples based on the representations obtained from self-supervised training~\cite{hacohen2022active,wang2021unsupervised}. These approaches select all labeled samples in a single shot. Hence, these methods can be easily employed in the context of semi-supervised learning. For example, \textbf{USL} proposes clustering in the self-supervised pre-training feature space and selecting samples at the density peaks of each cluster for labeling~\cite{wang2021unsupervised}. However, these methods often suffer from low accuracy in pseudo-labeling when the labeled samples are limited. In contrast, our method builds a more tightly coupled active self-semi-supervised learning framework and constructs an active learning strategy from the perspective of improving the accuracy of PPL.

\section{Revisiting Self-Supervised Pre-training Initialization for Semi-Supervised Training}
\label{sec:insight}

This section explores whether semi-supervised training can effectively harness valuable information acquired from self-supervised pre-training through weight initialization. Here, the framework of fine-tuning self-supervised pre-trained models using semi-supervised training methods is also referred to as SelfMatch~\cite{kim2021selfmatch}. Specifically, we initialize a semi-supervised model using weights pre-trained with the self-supervised method, Simsiam~\cite{chen2021exploring}. Subsequently, we conduct semi-supervised training using FlexMatch~\cite{zhang2021flexmatch}, one of the state-of-the-art semi-supervised learning algorithms. The semi-supervised training is executed on CIFAR-10~\cite{krizhevsky2009learning} with 10 and 40 labeled samples respectively, and on CIFAR-100~\cite{krizhevsky2009learning} with 200 and 400 labeled samples, where all the labeled samples are selected randomly. The network architecture and hyper-parameters for semi-supervised training align with the prior work~\cite{zhang2021flexmatch}. 

\begin{figure}[!htbp]
  \centering
  
  \begin{subfigure}{0.49\textwidth}
    \includegraphics[width=\linewidth]{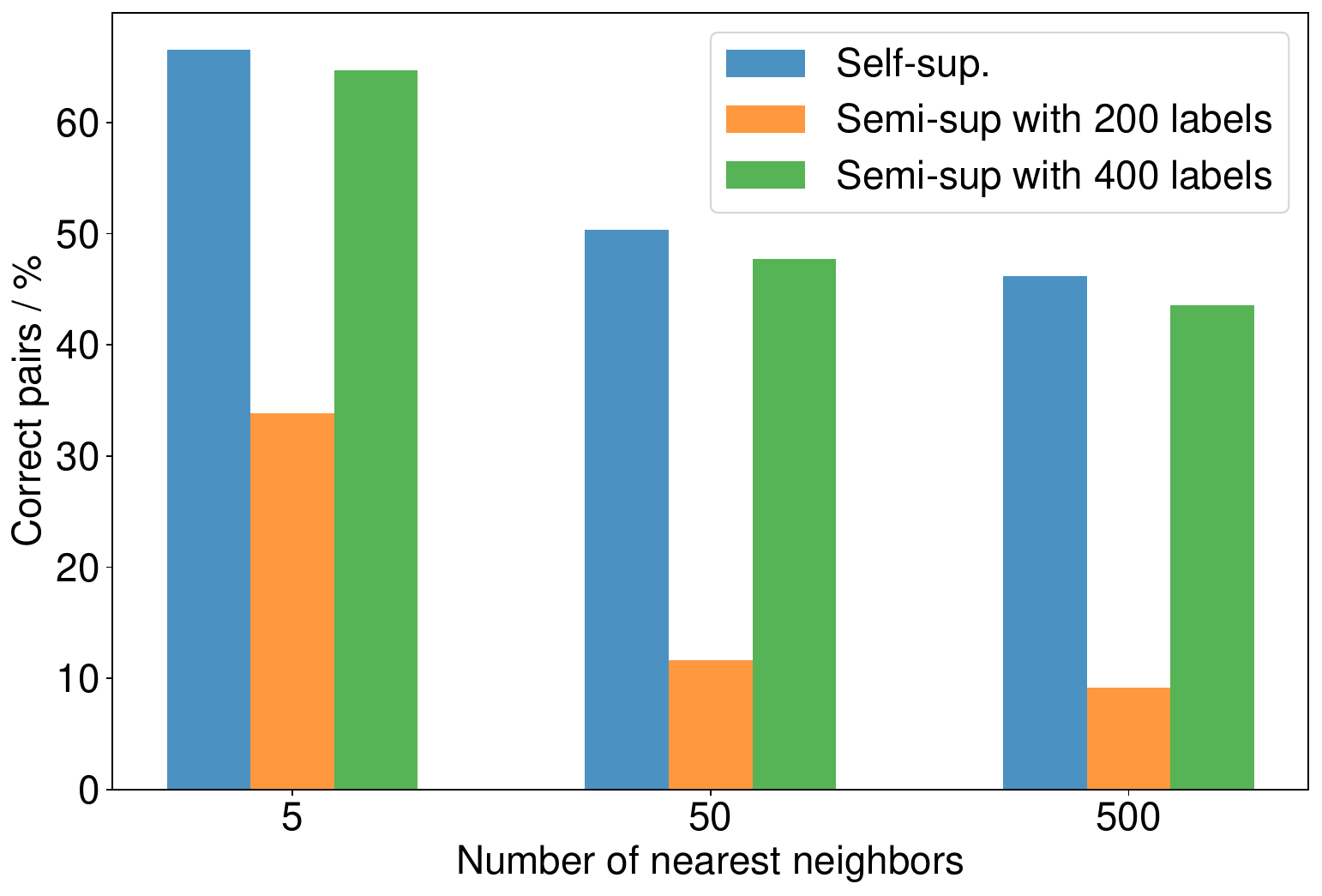}
    \caption{CIFAR-100}
    \label{fig:insight_c100}
  \end{subfigure}
  \begin{subfigure}{0.49\textwidth}
    \includegraphics[width=\linewidth]{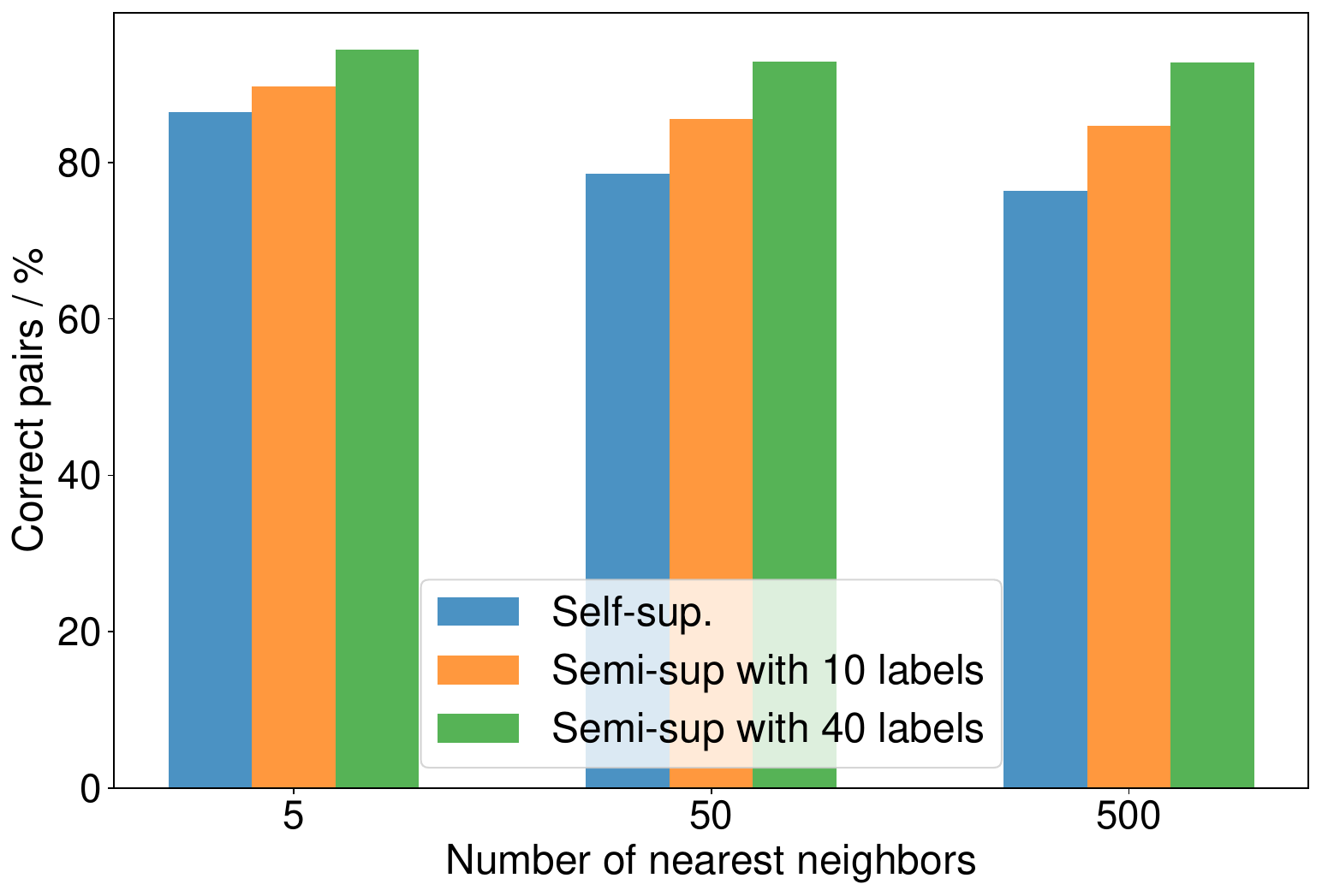}
    \caption{CIFAR-10}
    \label{fig:insight_c10}
  \end{subfigure}

  \caption{Consistency of sample labels with neighboring sample labels in self-supervised and semi-supervised feature spaces. The model was initialized with the self-supervised pre-training weights and then further trained using FlexMatch.}
  \label{fig:insight_feas}
\end{figure}

Self-supervised pre-training weight initialization allows downstream tasks to start training in a well-established pre-trained feature space~\cite{chen2020simple,grill2020bootstrap}, thereby improving the performance of downstream tasks. Therefore, the quality of features is a crucial indicator to assess whether semi-supervised training retains valuable information obtained from self-supervised training. Following~\cite{van2020scan}, we compute the degree of consistency between the label of the sample and the label of its nearest neighbors within the feature space. For the CIFAR-100 dataset, with few labeled samples, semi-supervised training undermines the valuable information obtained from self-supervised pre-training. As depicted in Fig.~\ref{fig:insight_c100}, the label consistency between samples and their nearest neighbor samples in the semi-supervised feature space notably declines compared to that in the self-supervised feature space. 



\begin{figure}[!htb]
\centering
\includegraphics[width=0.5\textwidth]{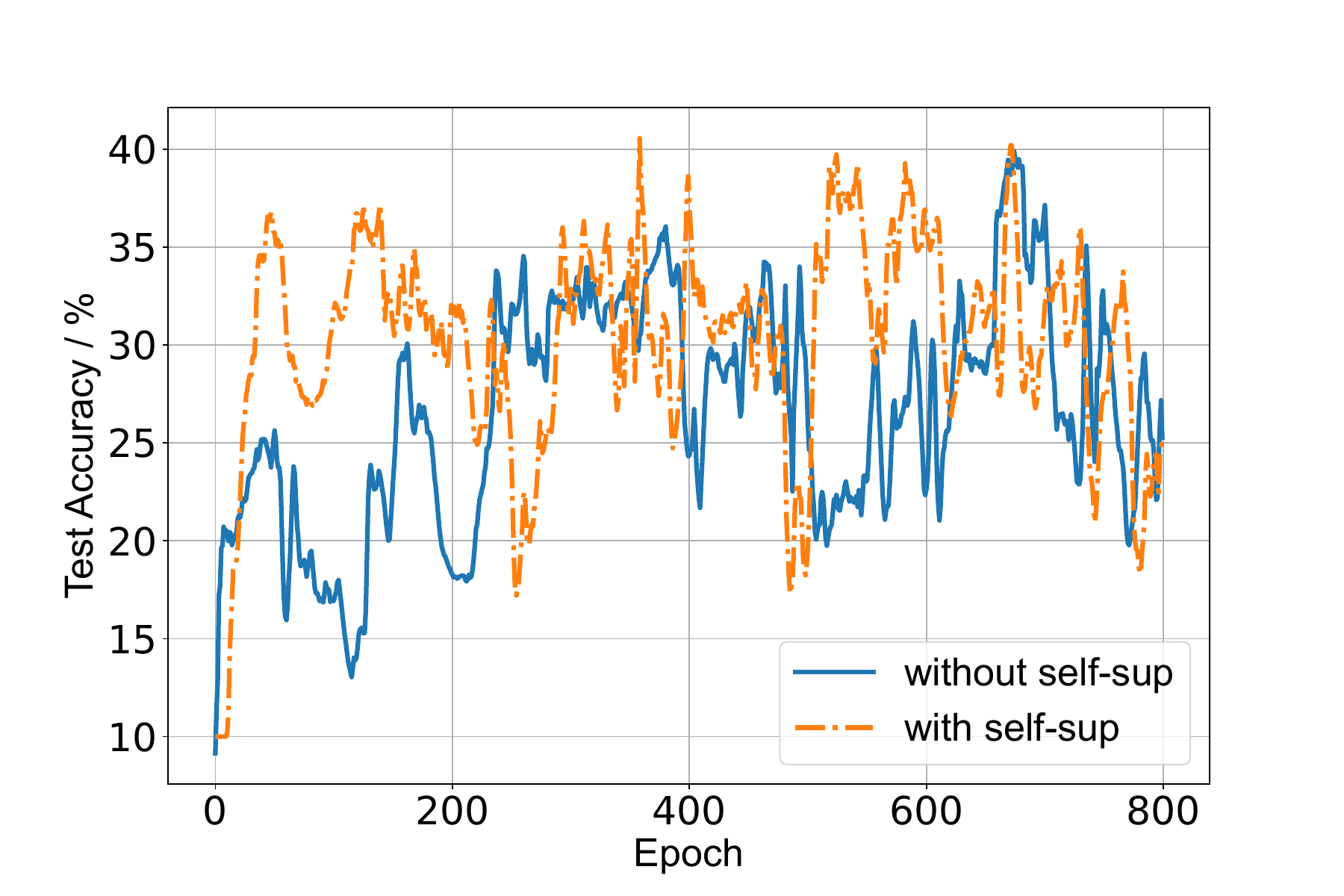}
\caption{ Test accuracy of the semi-supervised models initialized with self-supervised pre-training and random initialization. Specifically, semi-supervised training utilizing 10 labeled samples on CIFAR-10. Semi-supervised training method employed: FlexMatch.}
\label{fig:insight_c10_acc}
\end{figure}

For the simpler dataset, CIFAR-10, semi-supervised learning can generate features superior to those from self-supervised pre-training, as shown in Fig.~\ref{fig:insight_c10}. Yet, in scenarios with fewer annotations (10 labeled samples), semi-supervised training rarely benefits from self-supervised initialization. As illustrated in Fig.~\ref{fig:insight_c10_acc}, during the initial phase of semi-supervised training (approximately the first 200 epochs), self-supervised initialization enhances the performance of the semi-supervised model. However, as semi-supervised training progresses, the test accuracy of models initialized randomly and those using self-supervised pre-training initialization becomes nearly identical.

\begin{table}[!htb]
\small
\begin{center}
\caption{Impact of self-supervised pre-training weight initialization on semi-supervised performance. The experiments adopted the semi-supervised training method, FlexMatch. }
\label{table:insight_testacc}
\begin{tabular}{l|ll|ll}
\hline\noalign{}
Self-sup. & \multicolumn{2}{c|}{CIFAR-10} & \multicolumn{2}{c}{CIFAR-100} \\
Initialization & 10 labels & 40 labels & 200 labels & 400 labels \\
\hline
None & 59.06$\pm$19.80 & 94.86$\pm$0.05 & 30.59$\pm$1.69 & 46.11$\pm$2.83\\
Simsiam & 57.03$\pm$14.86 & 94.63$\pm$0.34 & 28.63$\pm$4.13 & 51.24$\pm$1.02 \\
\hline
\end{tabular}
\end{center}
\end{table}

Additionally, we demonstrated the impact of initializing with self-supervised pre-training weights on semi-supervised training performance, as shown in Table~\ref{table:insight_testacc}, indicating that it does not improve performance when labeled data is limited. In conclusion, weight initialization struggles to effectively transmit valuable information obtained from self-supervised training to the semi-supervised model. To address this issue, we propose to construct PPL, serving as an intermediary to facilitate semi-supervised models in fully leveraging the valuable information acquired from self-supervised pre-training.


\section{Method}
\label{sec:method}

We frame the problem and introduce the basic framework of semi-supervised learning in section~\ref{sec:prelim}. Subsequently, in sec.~\ref{sec:semisup_training}, we elaborate on leveraging PPL as an intermediary to facilitate the transfer of valuable information from self-supervised pre-training models to semi-supervised models. And sec.~\ref{sec:prior_pslabel} details the generation of PPL through label propagation on the self-supervised features, $f_{self}$. Finally, in sec.~\ref{sec:ss_active}, we propose an active learning strategy aimed at improving the accuracy of PPL.

\subsection{Preliminaries on Semi-Supervised Training}
\label{sec:prelim}
Semi-supervised learning trains a model, denoted as M($\cdot$), which comprises a feature extractor and a classification head. This training process leverages both a labeled dataset, $D_{l}=\{(x_{i},y_{i})\}_{i=1}^{N_1}$, and an unlabeled dataset, $D_{u}=\{(x_{i})\}_{i=N_1+1}^{N_1+N_2}$. The objective is to achieve superior performance compared to training exclusively on the labeled dataset. Typically, $N_2 \gg N_1$. The loss function for semi-supervised training, as depicted in Eq.~(\ref{eq:semisup_loss}), consists of two components: the cross-entropy loss on labeled samples, $L_{l}$, and the consistency loss on unlabeled samples, $L_{un}$. The hyper-parameter $\lambda$ governs the trade-off between these two losses. The $L_{l}$, as illustrated in Eq.~(\ref{eq:sup_loss}), represents the cross-entropy loss, CE, on weakly augmented labeled samples, where weak augmentation, $A_{w}$, follows the definition in previous research~\cite{sohn2020fixmatch}, incorporating operations such as image flipping.

\begin{equation}
    L = L_{l} + \lambda L_{un},
    \label{eq:semisup_loss}
\end{equation}

\begin{equation}
    L_{l} = \operatorname{CE}(\text{M}(A_{w}(x_{l})), y_{l}),
    \label{eq:sup_loss}
\end{equation}

The basic consistency loss, $L_{un}$, serves two purposes: using high-confidence model predictions as pseudo-labels to train unlabeled data and encouraging consistent predictions for inputs subjected to perturbations (image augmentations). As formulated in Eq.~(\ref{eq:con_loss0}), it involves applying both weak augmentation, $A_{w}$, and strong augmentation, $A_{s}$, to the unlabeled samples~\cite{sohn2020fixmatch}. Then, the model predicts unlabeled samples with weak augmentation and strong augmentation as $y_{pre,w}=\text{M}(A_{w}(x_{u}))$ and $y_{pre,s}=\text{M}(A_{s}(x_{u}))$, respectively. Here, $\mu$ is a ratio of the number of unlabeled samples and labeled samples in each training batch, and $B$ is batch size. $\mathbb{I}$ is an indicator function used to identify reliable predictions. When the confidence of its input surpasses the specified threshold $\tau$, the $\mathbb{I}$ returns 1; otherwise, it outputs 0. Additionally, the definition of $A_{s}$ aligns with that in prior research~\cite{sohn2020fixmatch}, including techniques such as cutout. 

\begin{equation}
    {{L_{un} = \dfrac{1}{\mu B} \sum_{b=1}^{\mu B} \mathbb{I}(\max(y_{pre,w})) \operatorname{CE}(y_{pre,s}, \arg \max(y_{pre,w})).
    \label{eq:con_loss0}}}
\end{equation}

\begin{figure*}[!ht]
\begin{center}
\includegraphics[width=0.95\textwidth]{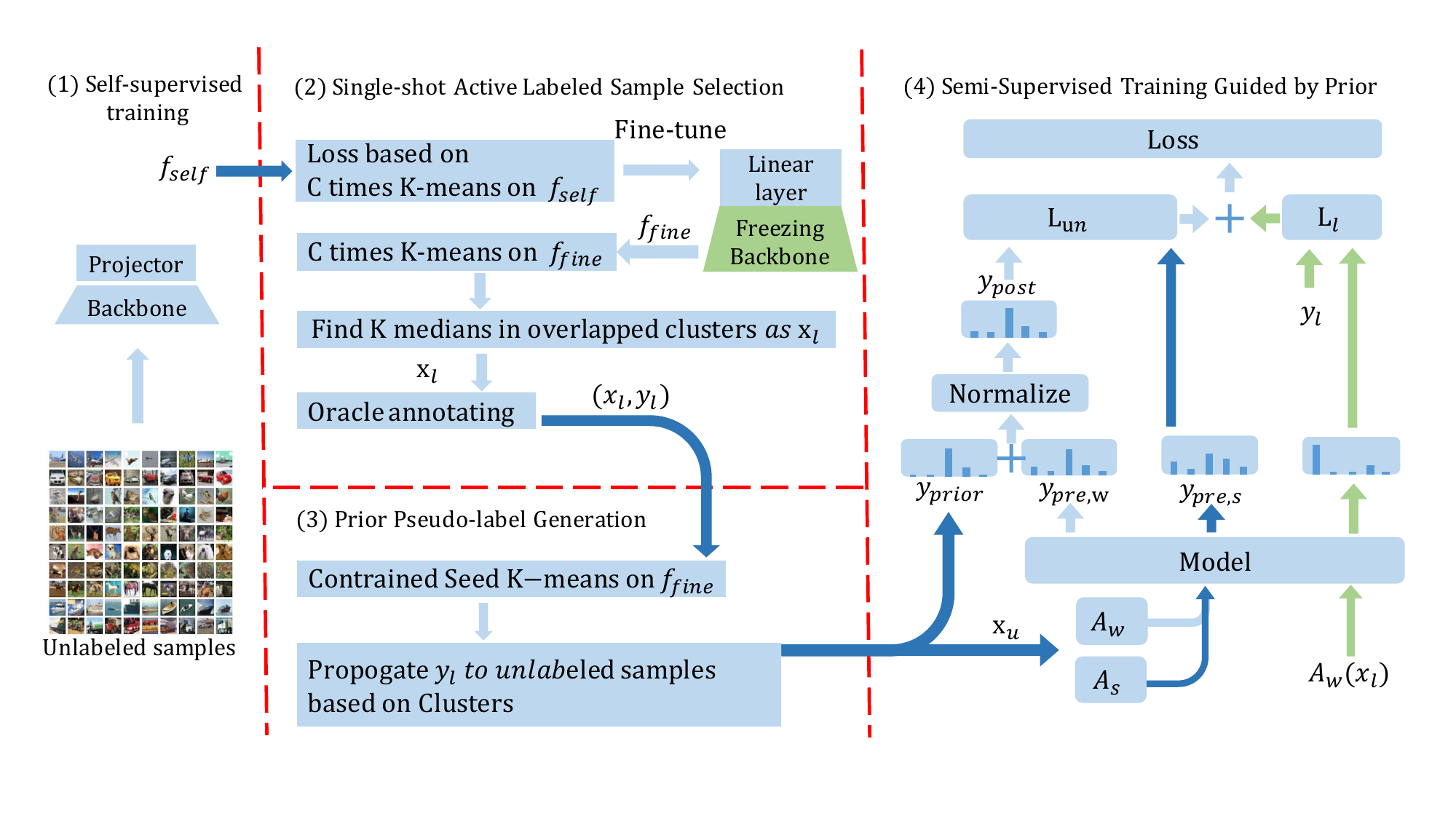}
\end{center}
\caption{The framework of our Active Self-Semi-Supervised learning(AS3L). AS3L consists of four components: (1) Obtaining self-supervised feature $f_{self}$ follows~\cite{chen2021exploring}; (2) Selecting labeled samples based on $f_{self}$ (Sec.~\ref{sec:ss_active}); (3) Label Propagation based on clusters to get PPL $y_{prior}$ (Sec.~\ref{sec:prior_pslabel}); (4) Semi-supervised training guided by $y_{prior}$ (Sec.~\ref{sec:semisup_training}).}
\label{fig:as4l}
\end{figure*}

\subsection{Active Self-Semi-Supervised Learning Framework}
\label{sec:semisup_training}

Existing methods typically initialize the model using weights pre-trained through self-supervised learning and then proceed with semi-supervised training to enhance performance. However, as demonstrated in sec.~\ref{sec:insight}, when the number of labeled samples is limited, valuable self-supervised features may be distorted during semi-supervised training, reducing the benefits of initialization with self-supervised pre-training weights. To better utilize self-supervised pre-training information and improve model performance, we propose extracting and transferring information from pre-trained features to semi-supervised training through pseudo-labels. This approach ensures that valuable information obtained from pre-training is preserved and effectively used, even as the model undergoes weight updates.

As depicted in Fig.~\ref{fig:as4l}, our framework leverages various techniques. Initially, the entire dataset $D=D_{l}\cup D_{u}$ undergoes self-supervised training to generate feature representations $f_{self}$. Our active learning strategy then selects samples and queries their labels to construct the labeled dataset $D_{l}$ based on $f_{self}$. After that, we propagate labels in the pre-trained feature space to obtain a group of pseudo-labels. These pseudo-labels serve as intermediaries to transfer valuable information from self-supervised pre-training to semi-supervised models. Given that these pseudo-labels are obtained before semi-supervised training, we refer to them as prior pseudo-labels (PPL), denoted as $y_{prior}$. The PPL is used for the unlabeled dataset $D_{u}=\{(x_{i}, y_{prior,i})\}_{i=N_1+1}^{N_1+N_{2}}$. Finally, a semi-supervised model is trained by combining these PPL with the model predictions, resulting in enhanced overall performance.

\begin{equation}
    {{L_{un} = \dfrac{1}{\mu B} \sum_{b=1}^{\mu B} \mathbb{I}( \max(y_{post}))\text{CE}(y_{pre,s}, \arg \max(y_{post})) }},
    \label{eq:con_loss}
\end{equation}

\begin{equation}
    y_{post} = \left\{ 
    \begin{array}{lcl}
    \operatorname{normalize}(y_{prior} + y_{pre,w}), & & {t \leq{T}}; \\
     y_{pre,w}, & & {t > T }. \\
    \end{array}
    \right.
    \label{eq:ypost}
\end{equation}

\textbf{Semi-Supervised Training Guided by PPL:} Assuming we have obtained prior pseudo-labels, $y_{prior}$, through label propagation on $f_{self}$ (which will be discussed in the next section), we integrate $y_{prior}$ into the existing semi-supervised training framework by re-formulating the consistency loss as Eq.~(\ref{eq:con_loss}). Instead of using the model prediction, $y_{pre,w}$, as the target for training unlabeled data, we combine $y_{prior}$ with $y_{pre,w}$ to generate a new training target, $y_{post}$, as given in Eq.~(\ref{eq:ypost}). Since this target is derived from both the semi-supervised model and the prior pseudo-labels, we slightly abuse the term and refer to it as posterior pseudo-labels.

In Eq.~(\ref{eq:ypost}), $T$ denotes a pre-defined switching point. This adjustment ensures that, during the early stages of semi-supervised training (i.e., before $T$ training iterations), when the model prediction is less accurate, $y_{prior}$ helps to guide the semi-supervised training. As the model predictions become more accurate than $y_{prior}$, the framework uses model predictions as pseudo-labels, resembling a conventional semi-supervised training approach. This transition helps mitigate the impact of PPL inaccuracies on the semi-supervised training. We empirically found that setting a rough switching point $T$ yields favorable results, as discussed in sec.~\ref{sec:switch_exp}.

\subsection{Prior Pseudo-label Generation}
\label{sec:prior_pslabel}
We generate PPL, $y_{prior}$, by applying label propagation based on clusters. Instead of using basic K-means, we employ constrained seed K-means~\cite{basu2002semi} to leverage labeled sample constraints, which enhances the quality of clustering by using label information when updating cluster centers. Then, the labels of the labeled samples are propagated to all unlabeled samples within the same cluster and then normalized to derive PPL. 

Another remaining issue is determining the value of $K$ for clustering. It is worth noting that as the number of clusters $K$ increases, the likelihood of samples within the same cluster sharing the same label also increases. Therefore, to improve the accuracy of PPL, it is advisable to increase $K$. However, as $K$ rises, the number of samples per cluster decreases, leading to an increase in the number of clusters devoid of any labeled samples. This is particularly evident when $K$ surpasses the number of labeled samples, resulting in a larger proportion of unlabeled samples not being associated with any labels. As a trade-off, we adopt a strategy of performing multiple clustering runs (denoted as $C$) and using different values of $K$ for each run. This approach ensures that the majority of unlabeled samples are encompassed by labeled samples, while those located farther away from any labeled samples exhibit lower confidence levels.

\subsection{Single-shot Active Labeled Sample Selection}
\label{sec:ss_active}

Our active learning strategy primarily focuses on improving the accuracy of PPL, which facilitates the transfer of valuable information from a self-supervised model to a semi-supervised model. For this purpose, existing active learning methods based on uncertainty and diversity principles are not suitable, as they tend to select challenging samples that exhibit high uncertainty~\cite{gao2020consistency} or are distant from the feature space of labeled samples~\cite{sener2018active}. These challenging samples often imply difficulty in differentiation within the self-supervised features, $f_{self}$. In other words, these samples are likely to have different labels from their neighbors (samples with similar features), leading to lower PPL accuracy when using these samples for label propagation. 

In contrast, selecting representative samples~\cite{settles2009active} is more appropriate for our scenario. Since our PPL is derived through clustering-based label propagation, in scenarios with limited annotations, the PPL for all samples within a cluster is determined by the labeled samples within the same cluster. In other words, a relatively small number of labeled samples play a crucial role in determining the PPL for the entire cluster. Therefore, we aim to select labeled samples that well represent the majority of samples within their cluster, ensuring that the selected labeled samples have consistent labels with most samples in their cluster.

Building upon the observations that within the self-supervised feature space, $f_{self}$, samples located in the same cluster tend to have similar labels, and those near the center of the cluster are more likely to share the same label as the majority of samples in that cluster (Sec. \ref{sec:cluster_self}). Our active strategy selects samples close to the center of the cluster to query labels. 

\begin{figure*}[!htbp]
  \centering
  
  \begin{subfigure}{0.43\textwidth}
    \includegraphics[width=\linewidth]{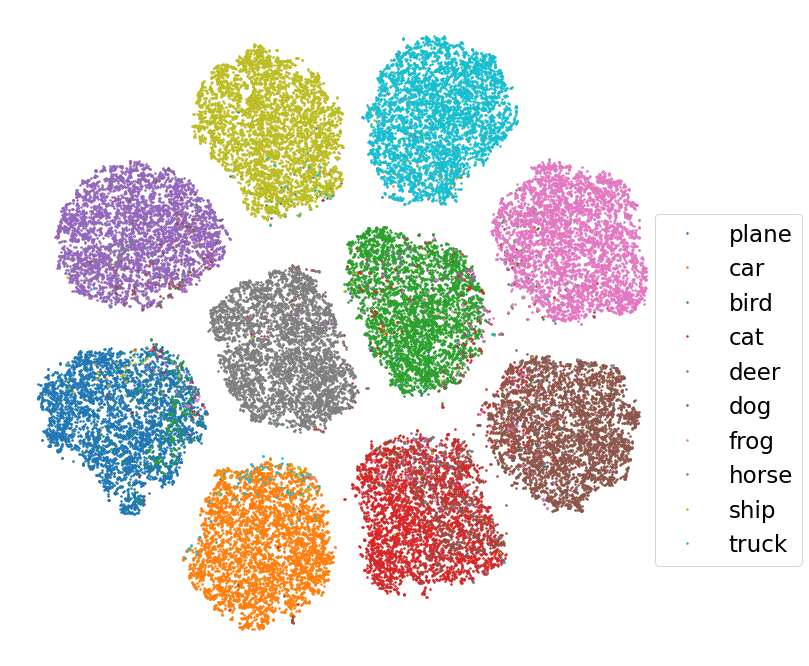}
    \caption{Semi-Supervised feature $f_{semi}$}
    \label{fig:semi-feas}
  \end{subfigure}
  \begin{subfigure}{0.43\textwidth}
    \includegraphics[width=\linewidth]{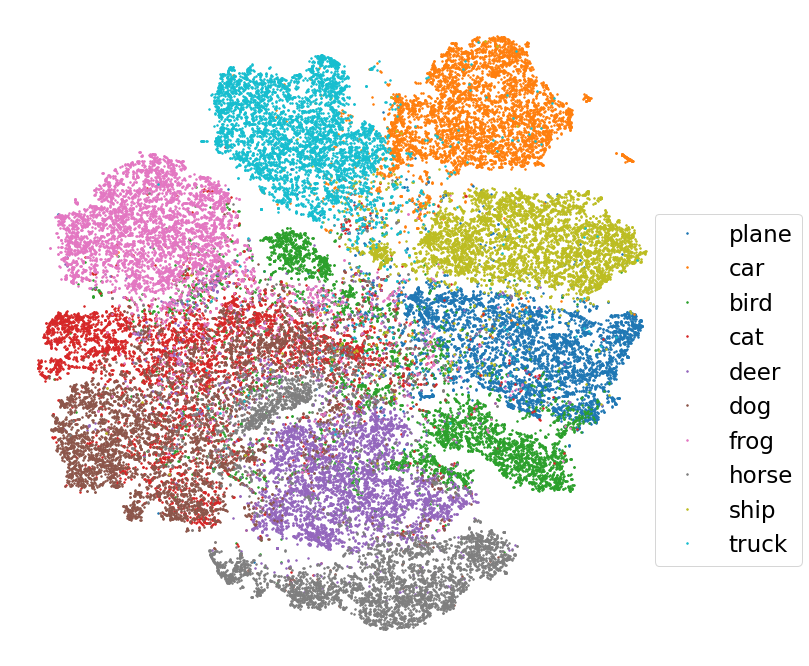}
    \caption{Self-Supervised feature $f_{self}$}
    \label{fig:self-feas}
  \end{subfigure}

  \caption{T-SNE visualization of semi-supervised and self-supervised features, where semi-supervised features are trained with 40 labels on CIFAR-10 followed our method. Self-supervised features seem to be more loosely clustered, and the boundaries between clusters are not as well defined.}
  \label{fig:feat_vis}
\end{figure*}

\textbf{Fine-tuning features:} While self-supervised training has been established as effective in producing high-quality features, $f_{self}$, with excellent performance in linear evaluation~\cite{chen2020simple,chen2021exploring}, we experimentally find that clustering directly on $f_{self}$ is not a good choice. One potential reason is that self-supervised features exhibit a distinct distribution compared to features trained with labeled data, as shown in Fig.~\ref{fig:feat_vis}. Self-supervised features tend to scatter due to their training on finer-grained proxy tasks, resulting in a larger distance between features of the same class and a smaller distance between features of different classes. This characteristic can impact the performance of the clustering algorithm and, potentially, the accuracy of our pseudo-labels. To achieve a more suitable feature space, we fine-tune these features based on multiple clusters before selecting samples for labeling.

To bring samples within the same cluster closer together, the mean squared error (MSE) loss is employed to force samples in the same cluster to be close to their centers. Also, to improve robustness, we perform $C$ runs of K-means on $f_{self}$, with the final loss as defined by Eq.~(\ref{eq:loss}), where $N$ is the number of samples within the whole dataset, $f_{fine,i}$ is $i_{}^{th}$ sample's fine-tuned feature, $f_{center,i,j}$ denotes the cluster center assigned to $i_{}^{th}$ sample during the $j_{}^{th}$ clustering run, where $j$ refers to the specific iteration out of a total of $C$ runs of K-means clustering. During the fine-tuning process, as the loss is defined based on $C$ times clustering with randomness, those samples that are stable within the same cluster become closer, while those attracted by different cluster centers do not approach any specific center, thereby enhancing the clustering results.

\begin{equation}
Loss = \dfrac{1}{ CN } \sum_{i}\sum_{j}\Vert f_{fine,i} - f_{center,i, j} \Vert_{2}^{2}.
\label{eq:loss}
\end{equation}

Additionally, considering the computational cost while still aiming to retain the approximate self-supervised feature structure, we add a single linear layer to the self-supervised trained encoder. During training, the encoder weights are frozen and only the weights of the newly added linear layer are updated. To reduce the computational cost, we pre-compute the self-supervised features for all samples. This allows us to directly use the $f_{self}$ as input during the training of the new linear layer, eliminating the need to forward-pass through the backbone in each training iteration. Finally, we select labeled samples and generate PPL based on the output of the linear layer, denoted as $f_{fine}$.

\textbf{Select Labeled Samples:} To improve robustness, conduct $C$ rounds of K-means clustering on $f_{fine}$. Subsequently, identify samples consistently assigned to the same class across all $C$ clustering rounds. Calculate the mean of the features for these samples to represent the center of that class. Then, select and annotate as the labeled set the samples that are closest to these $K$ class centers.

\section{Experiments}
\label{sec:res}

\subsection{Results on CIFAR-10, CIFAR-100 and STL-10}
Our method is first evaluated on the common benchmark: CIFAR-10~\cite{krizhevsky2009learning}, CIFAR-100~\cite{krizhevsky2009learning}, STL-10~\cite{coates2011analysis}. Both CIFAR-10 and CIFAR-100 contain 50,000 training samples and 10,000 test samples, all with image size of 32$\times$32. STL-10 contains 5,000 labeled training samples, 100,000 unlabeled training samples and 8,000 test samples. The resolution of all samples is 96$\times$96. We experimented with labeled sample sets of various sizes, especially with fewer annotation samples than in previous papers (10 labeled samples for CIFAR-10, 200 labeled samples for  CIFAR-100, and 20 labeled samples for STL-10).
\subsubsection{Baseline Methods}
We compare our method with (1) Semi-supervised learning from scratch methods: FixMatch~\cite{sohn2020fixmatch}, FlexMatch~\cite{zhang2021flexmatch} and CoMatch~\cite{li2021comatch}. (2) Semi-supervised learning initialized with the self-supervised model: SelfMatch~\cite{kim2021selfmatch}. SelfMatch refers to a training framework that utilizes standard semi-supervised training methods to fine-tune models initialized with self-supervised pre-training weights. This framework applies to various self-supervised pre-training and semi-supervised training methods. To set a fair baseline, we employ identical semi-supervised (FlexMatch) and self-supervised training (SimSiam~\cite{chen2021exploring}) methods for SelfMatch as those utilized in our framework. (3) Plug-in model selection method, Sel~\cite{kim2022propagation}, that is specially optimized for semi-supervised training with few annotations. (4) Constructing self-supervised signals to enhance semi-supervised training in scenarios with limited annotations, Less~\cite{lucas2022barely}. The annotated samples for the aforementioned baseline methods were randomly selected. (5) Active semi-supervised learning method: USL~\cite{wang2021unsupervised} conducts semi-supervised training by actively selecting labeled samples. To establish a fair baseline, we utilize the same self-supervised pre-training weight initialization for our method, SelfMatch, and USL.

\subsubsection{Implemention details}
\label{sec:setup}

Many existing semi-supervised learning methods involve the use of pseudo-labeling techniques. Our method readily integrates with these techniques. In our implementation, we conduct experiments with FlexMatch, one of the state-of-the-art semi-supervised approaches. For the self-supervised pre-training, any self-supervised method can be employed. For this study, we chose SimSiam~\cite{chen2021exploring}, a state-of-the-art self-supervised method, because it has demonstrated good pre-training performance even on small-scale datasets like CIFAR-10. 
The same backbone as used in the semi-supervised learning stage is utilized in the self-supervised phase. We adhere to the hyper-parameter settings proposed by SimSiam~\cite{chen2021exploring}. The network weights obtained from self-supervised learning serve as the initialization for the backbone of the semi-supervised model. 

We set network architectures following previous work \cite{zhang2021flexmatch}: WRN-28-2~\cite{zagoruyko2016wide} for CIFAR-10, WRN-28-8 for CIFAR-100 and WRN-37-2 for STL-10. For CIFAR-10, CIFAR-100 and STL-10. We set hyper-parameters following TorchSSL~\cite{zhang2021flexmatch}: SGD with momentum 0.9, initial learning rate 0.03, $\lambda$ is 1, $\mu$ is 7, the batch size is 64, total training iterations is $1024\times1024$ and a cosine annealing learning rate scheduler. 

The hyper-parameters involved in the active learning and label propagation in this paper include the number of times clustering was performed, $C$, and the number of classes in each clustering, $K$. Specifically, clustering is performed 6 times ($C$ is set to 6) in both the active sampling strategy and label propagation. For active sampling, the number of classes in each clustering, $K$, is equal to the number of selected samples. In label propagation, clustering is accomplished using Constrained Seed K-means~\cite{basu2002semi}. The minimum value of $K$ is set to match the number of categories in the dataset. For CIFAR-10 and STL-10, $K$ takes values of 10, 20, 30, 40, 50, and 60. For CIFAR-100, $K$ is set to 100, 200, 300, 400, 500, and 600, respectively. The linear layer mentioned in Sec.~\ref{sec:ss_active} has the same dimension as the final layer of the backbone, and fine-tuning features are trained for 40 epochs. Additionally, the $T$ for switching from PPL-guided pseudo-labels to regular pseudo-labels during semi-supervised training is set to the 60th epoch.

\subsubsection{Results}
\label{sec:main_results1}

\begin{table*}[!hbt]
\caption{Comparison of accuracy on CIFAR-10, CIFAR-100 and STL-10. All results are averaged over 3 runs. The best results are shown in red and the second best results are shown in blue.}
\centering

\label{table:main result}

\resizebox{\linewidth}{!}{
  
\begin{tabular}{l|ll|ll|ll|ll}
\hline\noalign{}
 & Active & Self-sup. & \multicolumn{2}{c|}{CIFAR-10} & \multicolumn{2}{c|}{CIFAR-100} & \multicolumn{2}{c}{STL-10} \\
  & Learning & Initialization &  10 labels & 40 labels & 200 labels & 400 labels & 20 labels & 40 labels \\
\hline
Fully-Supervised  & - & - & \multicolumn{2}{c|}{95.38} & \multicolumn{2}{c|}{80.70} & \multicolumn{2}{c}{-} \\
\hline
FixMatch & No & None & 60.01$\pm$7.41 & 85.57$\pm$5.21 & 35.83$\pm$1.63 & 46.04$\pm$1.41 & 43.24$\pm$6.32 & 60.92$\pm$5.60 \\
FixMatch-Sel & No & None & 65.73$\pm$10.32 & 89.87$\pm$4.96 & 35.78$\pm$1.58 & 46.05$\pm$1.28 & 45.45$\pm$3.24 & 63.93$\pm$9.65 \\
FlexMatch & No & None & 59.06$\pm$19.80 & \color[rgb]{1,0,0}{94.86$\pm$0.05} & 30.59$\pm$1.69 & 46.11$\pm$2.83 & 38.31$\pm$16.69 & 54.90$\pm$8.06 \\
CoMatch & No & None & 65.10$\pm$7.81 & 92.16$\pm$4.97 & 32.51$\pm$1.15 & 41.72$\pm$2.04 & - & -\\
LESS & No & None & 64.40$\pm$10.90 & 93.20$\pm$2.10 & 42.50$\pm$3.20 & 51.30$\pm$2.40 & 48.98$\pm$5.19 & \color[rgb]{0,0,1}{64.20$\pm$5.10} \\

SelfMatch
(w FlexMatch) & No & \color[rgb]{1,0,0}{Simsiam} & 57.03$\pm$14.86 & 94.63$\pm$0.34 & 28.63$\pm$4.13 & 51.24$\pm$1.02 & 44.78$\pm$1.98 & 57.06$\pm$5.99 \\
SelfMatch-Sel(w FlexMatch) & No & \color[rgb]{1,0,0}{Simsiam} & 59.13$\pm$15.43 & \color[rgb]{0,0,1}{94.79$\pm$0.07} & 36.58$\pm$3.31 & 53.05$\pm$1.85 & 45.73$\pm$4.11 & 63.14$\pm$9.47\\
USL-Sel (w FlexMatch) & \color[rgb]{1,0,0}{Yes} & \color[rgb]{1,0,0}{Simsiam} & 55.76$\pm$6.45 & 89.22$\pm$0.79 & 25.31$\pm$10.32 & 29.73$\pm$4.77 & \color[rgb]{1,0,0}{55.90$\pm$1.33} & 61.94$\pm$5.14 \\
\hline

Ours (w FlexMatch) & \color[rgb]{1,0,0}{Yes} & \color[rgb]{1,0,0}{Simsiam} & 69.08$\pm$5.32 & 94.35$\pm$0.20 & \color[rgb]{0,0,1}{46.87$\pm$2.47} & 47.37$\pm$6.66 & 49.57$\pm$9.32 & 57.57$\pm$9.55 \\
Ours-Sel (w FlexMatch) & \color[rgb]{1,0,0}{Yes} & \color[rgb]{1,0,0}{Simsiam} & \color[rgb]{1,0,0}{83.80$\pm$10.57} & 94.32$\pm$0.09 & \color[rgb]{1,0,0}{48.40$\pm$5.78} & \color[rgb]{0,0,1}{59.16$\pm$2.37} & \color[rgb]{0,0,1}{51.11$\pm$6.88} & 61.89$\pm$7.73 \\
Ours-Early-Sel (w FlexMatch) & \color[rgb]{1,0,0}{Yes} & \color[rgb]{1,0,0}{Simsiam} & \color[rgb]{0,0,1}{82.17$\pm$11.09} & 93.72$\pm$0.22 & 44.80$\pm$5.07 & \color[rgb]{1,0,0}{61.50$\pm$2.00} & 46.32$\pm$7.77 & \color[rgb]{1,0,0}{65.15$\pm$2.44} \\
\hline
\end{tabular}
}
\end{table*}

The experimental results are shown in Table~\ref{table:main result}. Given the lack of labeled data for semi-supervised training, selecting the best model by validation set is infeasible. Following previous work \cite{kim2022propagation}, we report the median accuracy for the last 20 checkpoints. Notably, “Sel” denotes the use of the metric proposed by Kim et al.~\cite{kim2022propagation} to select the final model for evaluation. Our results are presented in 2 ways: first, the median accuracy of the last 20 models as “Ours”  and second, the accuracy of the models selected using “Sel”~\cite{kim2022propagation} as “Ours-Sel”, where the total training iteration is 1024$\times$1024. Additionally, to demonstrate the accelerated convergence of our method, we report the accuracy achieved with only 1/4 of the total training iterations, termed as “Ours-Early-Sel”.

\textbf{Effectiveness:} Our approach demonstrates significant improvements over most baselines, except for LESS, which shows similar performance under some experimental setups. Notably, the results with LESS were obtained under somewhat unfair conditions. LESS selects labeled samples through random stratified sampling (i.e., randomly selecting an equal number of samples per class), ensuring that all classes are covered even with limited annotations, which requires additional knowledge about the dataset. In contrast, our method selects samples without prior knowledge of the dataset's classes, contributing to the similar results observed between LESS and our method.

A comparison between FlexMatch and SelfMatch (with FlexMatch) indicates that solely relying on self-supervised pre-training initialization does not significantly improve the performance of semi-supervised training in nearly half of the experiments. Our method, incorporating initialization, PPL, and annotated sample selection, enables a more comprehensive utilization of valuable information from self-supervised pre-training, resulting in superior overall performance.

Furthermore, our method typically brings more improvements when dealing with a limited number of annotations (approximately 1 or 2 labels per class on average). As the number of labels increases, the benefits of our method decrease, especially for simple datasets. This is because semi-supervised training can achieve impressive performance with limited annotated data, reducing the necessity for additional assistance from PPL. For example, on CIFAR-10, the baseline method, FlexMatch, using only 40 annotated samples, can achieve accuracy comparable to fully supervised training, making the impact of PPL less prominent.

\textbf{Fast convergence:} Our method demonstrates accelerated convergence by explicitly employing PPL to initiate semi-supervised training. As shown in Table~\ref{table:main result}, our method (Ours-Early-Sel) achieves about 97\% accuracy with only 1/4 of the training iterations. It also outperforms baseline methods in most cases with just 1/4 of the training iterations. Additionally, we compared the computational time required for our method to reach the final accuracy of SelfMatch (w FlexMatch), which is initialized with self-supervised pre-trained weights. As illustrated in Table~\ref{table: running time}, our method achieves the final accuracy of SelfMatch in approximately 1/3 of the training time.

\begin{table}[!htb]
\small
\begin{center}
\caption{Comparison of practical running time on a single RTX 3090 GPU. SelfMatch time refers to the total runtime of complete semi-supervised training, while our method’s time indicates the duration required to reach SelfMatch’s final accuracy.}
\label{table: running time}
\begin{tabular}{l|ll}
\hline\noalign{}
Dataset & CIFAR-10 & CIFAR-100 \\
\hline
SelfMatch (w FlexMatch) & 98.82 (hrs) & 353.69 (hrs) \\
Ours (w FlexMatch) & 35.21 (hrs) & 121.04 (hrs)\\
\hline
\end{tabular}
\end{center}
\end{table}

\subsection{Results on ImageNet}
We conducted experiments on the large-scale dataset, ImageNet-1k~\cite{deng2009imagenet}, which has 1.28 million training images and 50,000 validation images. Following previous works~\cite{zhang2021flexmatch,wang2022debiased}, we adopted ResNet-50~\cite{he2016deep} as backbone. Due to computational resource limitations, we set hyper-parameters following the method of training based on self-supervised pre-training weights~\cite{wang2022debiased, cai2021exponential} instead of the original FlexMatch~\cite{zhang2021flexmatch}. The total training epoch is 50, with a change point $T$ at the 4th epoch.

FlexMatch and SelfMatch served as our baseline methods. We implemented two versions of the SelfMatch, employing different semi-supervised training approaches: FlexMatch and FixMatch. For FixMatch, we employed a variant known as FixMatch-EMAN, where EMAN~\cite{cai2021exponential} is utilized to enhance both pre-training and semi-supervised training effectiveness. To ensure fairness, both SelfMatch and our method are initialized with the same pre-trained weights, as shown in Table~\ref{table:imagenet}. They are respectively pre-trained using BYOL~\cite{grill2020bootstrap} and BYOL-EMAN~\cite{cai2021exponential}. The annotated samples for baseline methods were randomly selected, with a difference from existing semi-supervised learning experimental setups. In contrast to the common practice of randomly selecting $N$ labeled samples per class, our experiments were conducted in a more realistic setting. Specifically, we randomly selected annotated samples from the entire dataset.

The same trend is observed in experiments on ImageNet, as detailed in Table~\ref{table:imagenet}. Our method outperforms SelfMatch significantly, especially when labeled data is limited. This suggests that our method can more effectively leverage pre-trained models to enhance semi-supervised learning performance. Additionally, this improvement is consistent across different semi-supervised training methods, including FlexMatch and FixMatch-EMAN.

\begin{table}
\small
\begin{center}
\caption{Comparison of accuracy on ImageNet. The best results are shown in red and the second best results are shown in blue.}
\label{table:imagenet}
\begin{tabular}{lllllll}
\hline\noalign{}
Semi-Supervised & Self-Supervised & \multicolumn{2}{c}{0.2\% labels} & \multicolumn{2}{c}{1\% labels} \\
Method & Initialization & Top-1 & Top-5 & Top-1 & Top-5 \\
\hline
Fine-tune & BYOL & 26.0 & 48.7 & 53.2 & 68.8 \\
FlexMatch & None & 3.9 & 9.6 & 19.6 & 37.9 \\
SelfMatch (w FlexMatch) & BYOL & 30.3 & 48.8 & 57.3 & 80.1 \\
SelfMatch (w FixMatch-EMAN) & BYOL-EMAN & 32.7 & 52.7 & \color[rgb]{0,0,1}{59.6} & \color[rgb]{1,0,0}{81.6} \\
\hline
Ours (w FlexMatch) & BYOL & \color[rgb]{0,0,1}{37.3} & \color[rgb]{0,0,1}{55.9} & 58.3 & 79.9 \\
Ours (w FixMatch-EMAN) & BYOL-EMAN & \color[rgb]{1,0,0}{40.4} & \color[rgb]{1,0,0}{60.7} & \color[rgb]{1,0,0}{60.4} & \color[rgb]{0,0,1}{81.5} \\
\hline
\end{tabular}
\end{center}
\end{table}

\subsection{Computational Complexity}

The additional computational complexity of the proposed method can be divided into two main components: the active learning part and the PPL-guided semi-supervised training. The PPL-guided semi-supervised training introduces one additional plus and normalization operation per iteration compared to standard semi-supervised training. This part has a similar computational complexity to that of standard semi-supervised learning methods.

The computational complexity in active learning and PPL generation part is influenced by three sets of K-means clustering and feature fine-tuning. The first and second sets of K-means are performed on self-supervised features and fine-tuned features, respectively, for sample selection. The third set of K-means is applied to fine-tuned features for label propagation to generate prior pseudo-labels. Let $d$ be the feature dimension, $I$ the maximum number of iterations for K-means, and $N$ the number of samples in the dataset. Each set of K-means clustering is executed $C$ times (details are in Section~\ref{sec:setup}). The total computational complexity for clustering is $O(CUdI)$. For fine-tuning features, since only a simple linear network is trained rather than the entire neural network, the training time required is considerably less compared to semi-supervised training.

Overall, compared to standard semi-supervised learning methods, which require extensive training iterations of the entire neural network, the proposed method adds marginal additional computational time for active learning and sample selection.

\subsection{Model Detailed Analysis}
We further analyze our approach from six aspects: ablation studies for the whole framework, influence of switching point, comparisons of active learning strategies, ablation studies for our active sampling strategy, and the detailed analysis of pseudo-label propagation and clustering results on self-supervised features.

\subsubsection{Ablation Experiment of Framework}

Ablation experiments were performed in the CIFAR-10 with 10 labels. We maintained the same hyper-parameters, except for reducing the total training iterations to 112$\times$2304, equivalent to 2304 epochs. Two components of our method are evaluated: active labeled set selection and PPL. The results are shown in Table~\ref{table: ablation}. 
(1) When working with a limited number of labels, the active learning component yields more stable and accurate results in semi-supervised training.
(2) Whether for randomly selected or actively chosen annotated samples, the use of PPL significantly improves the performance of semi-supervised learning.

\begin{table}
\small
\begin{center}
\caption{Ablation Study on CIFAR-10. Accuracy is the average over 3 runs.}
\label{table: ablation}
\begin{tabular}{ll}
\hline
Ablation & Accuracy(avg. $\pm$ std.)\\ 
\hline
Random & 58.17 $\pm$ 13.01 \\ 
Random + PPL & 68.01 $\pm$ 15.96 \\ 
Active & 76.65 $\pm$ 5.29 \\
Active + PPL & 84.43 $\pm$ 5.19  \\
\hline
\end{tabular}
\end{center}
\end{table}

\subsubsection{Influence of switching point}
\label{sec:switch_exp}

\begin{figure}[!htbp]
  \centering
  
  \begin{subfigure}{0.49\textwidth}
    \includegraphics[width=\linewidth]{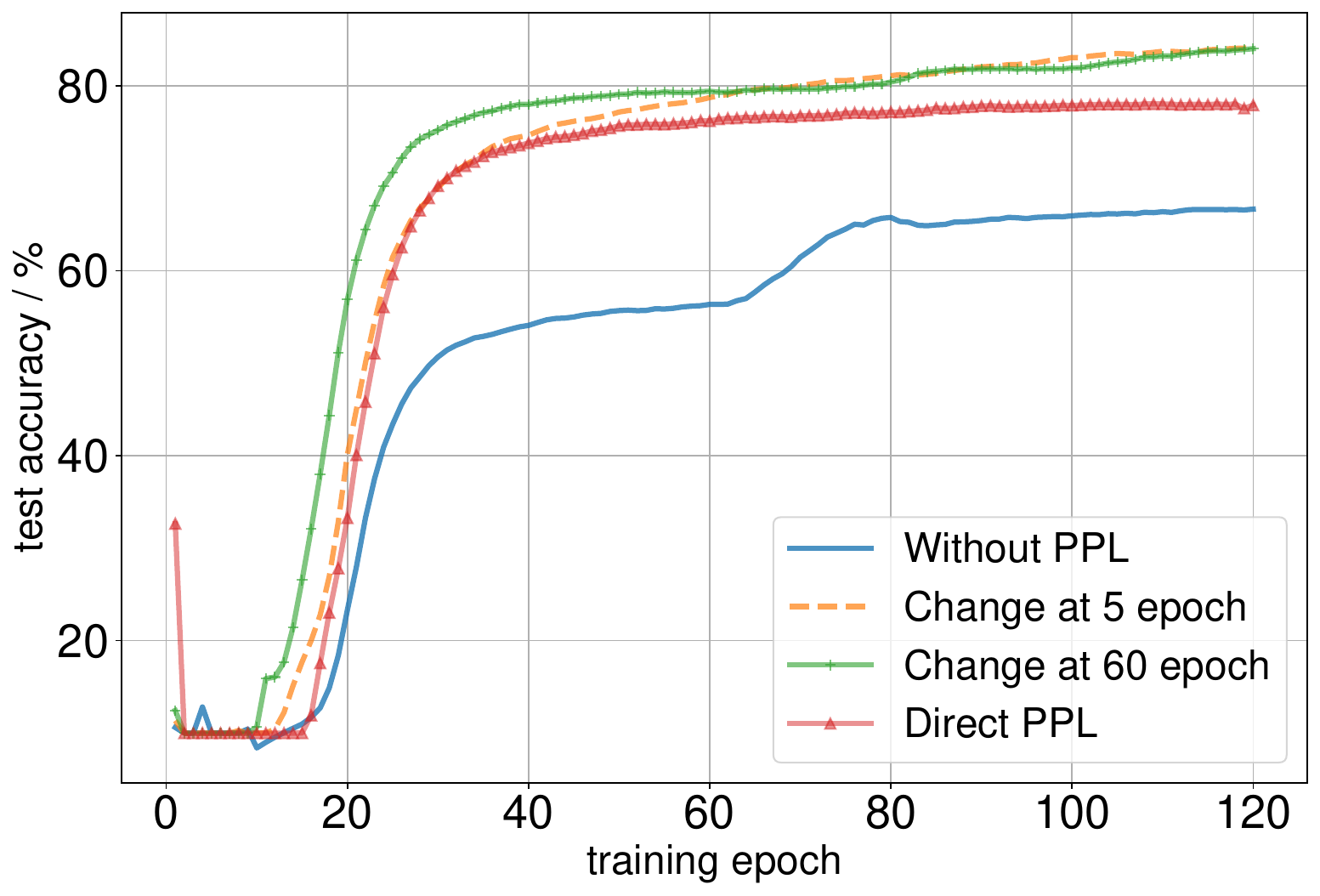}
    \caption{10 labels}
    \label{fig:switch_c10_lbl10}
  \end{subfigure}
  \begin{subfigure}{0.49\textwidth}
    \includegraphics[width=\linewidth]{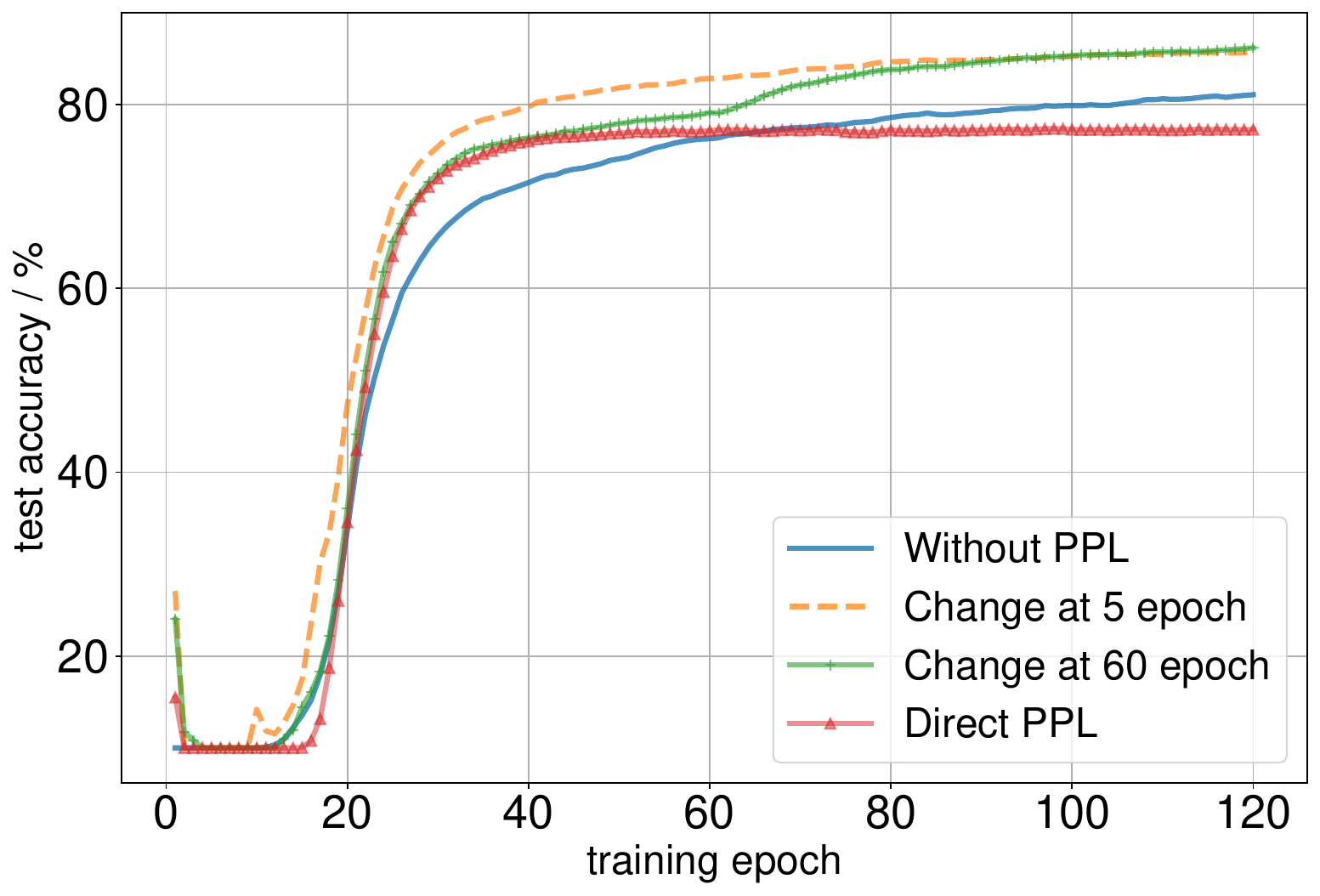}
    \caption{40 labels}
    \label{fig:switch_c10_lbl40}
  \end{subfigure}

  \caption{Test accuracy of semi-supervised training in CIFAR-10. In all experiments, the same self-supervised pre-training weight initialization is applied. Without PPL indicates the absence of PPL. Change at 5 epoch and change at 60 epoch refer to the utilization of PPL according to Eq.~(\ref{eq:ypost}), with $T$ as 5 epoch and 60 epoch, respectively. Direct PPL signifies using PPL in all training iterations.}
  \label{fig:switch_c10}
\end{figure}

To assess the impact of the PPL switch point, $T$, on semi-supervised training, we conducted four experiments on the CIFAR-10 dataset with 10 and 40 labeled samples, respectively. The experiments included: (1) Without PPL, (2) Employing PPL as in Eq.~(\ref{eq:ypost}) with $T$ set to 5 epochs, (3) $T$ set to 60 epochs, and (4) Direct PPL, i.e., $T$ equal to the total number of training iterations. The results are depicted in Fig.~\ref{fig:switch_c10_lbl10} and Fig.~\ref{fig:switch_c10_lbl40}.



PPL enhances semi-supervised training accuracy in the early stages, but continuous use of PPL (experiment setting (4)) hinders the sustained update of pseudo-labels, thereby impacting the final semi-supervised performance. For instance, in Fig.~\ref{fig:switch_c10_lbl40}, the semi-supervised training accuracy of experiment setting (1) surpasses that with continuous PPL (setting (4)) after training epochs exceed about 70. Hence, a switching mechanism, as represented by Eq.~(\ref{eq:ypost}), is necessary for guiding the semi-supervised training process. 

The specific placement of the switch point minimally affects the ultimate performance of semi-supervised training. Although setting a larger $T$ (i.e., using PPL in more training iterations) yields higher accuracy in the early stages of semi-supervised training with limited labeled data as shown in Fig.~\ref{fig:switch_c10_lbl10}, and setting a smaller $T$ helps avoid the constraint imposed by fixed PPL on the process of updating pseudo-labels in scenarios with more labeled data as shown in Fig.~\ref{fig:switch_c10_lbl40}, after the switch at point $T$, the semi-supervised training accuracy becomes consistent quickly for different $T$ settings. Therefore, the impact of setting a roughly appropriate switch point on the final performance of semi-supervised training is minimal.

\subsubsection{Ablation Study of Our Active Sampling Strategy}

We conducted ablation experiments to compare the effects of various components in our proposed active learning strategy on CIFAR-10. In these experiments, “K-medoids” refers to clustering only once, while “multi-clustering” involves clustering 6 times, as described in Sec.~\ref{sec:setup}. 

As shown in Table~\ref{table: ablation of AL}, fine-tuning the features results in significant improvements, providing better class coverage and more accurate pseudo-labels. Especially for the case with fewer annotations, where multi-clustering and fine-tuning features yield greater benefits.

\begin{table}
\small
\begin{center}
\caption{Ablation study of proposed active learning strategy on CIFAR-10. Accuracy reported is the average over 3 runs.}
\label{table: ablation of AL}
\begin{tabular}{lllll}
\hline
\quad & \multicolumn{2}{c}{Accuracy of $y_{prior}$} & \multicolumn{2}{c}{Class Coverage} \\
\# Labels & 10 & 40 & 10 & 40 \\
\hline\noalign{\smallskip}
K-medoids on $f_{self}$ & 44.90$\pm$1.88 & 69.62$\pm$0.83 & 7.2$\pm$1.4 & 10.0$\pm$0.0 \\
Multi-cluster on $f_{self}$ & 61.94$\pm$5.42 & 72.54$\pm$1.42 & 9.0$\pm$0.7 & 10.0$\pm$0.0\\
K-medoids on $f_{fine}$ & 64.70$\pm$6.27 & 72.72$\pm$2.77 & 9.2$\pm$0.8 & 10.0$\pm$0.0\\
Multi-cluster on $f_{fine}$ & \color[rgb]{1,0,0}{71.41$\pm$4.10} & \color[rgb]{1,0,0}{74.91$\pm$1.66} & \color[rgb]{1,0,0}{9.5$\pm$0.5} & 10.0$\pm$0.0\\
\hline
\end{tabular}
\end{center}
\end{table}

\begin{figure}[!htbp]
  \centering
  
  \begin{subfigure}{0.49\textwidth}
    \includegraphics[width=\linewidth]{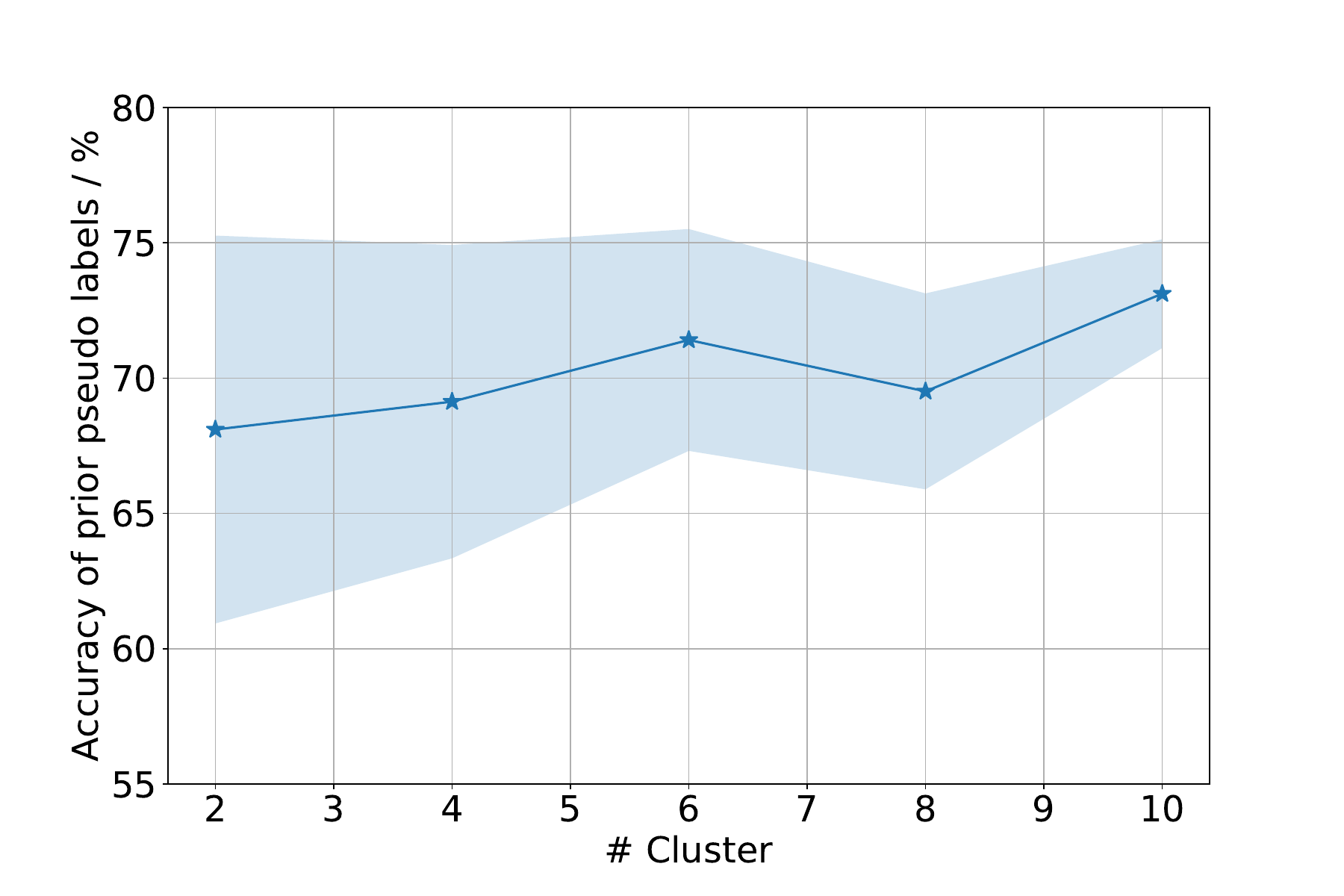}
    \caption{Accuracy of prior Pseudo-labels}
    \label{fig:prior acc hyper}
  \end{subfigure}
  \begin{subfigure}{0.49\textwidth}
    \includegraphics[width=\linewidth]{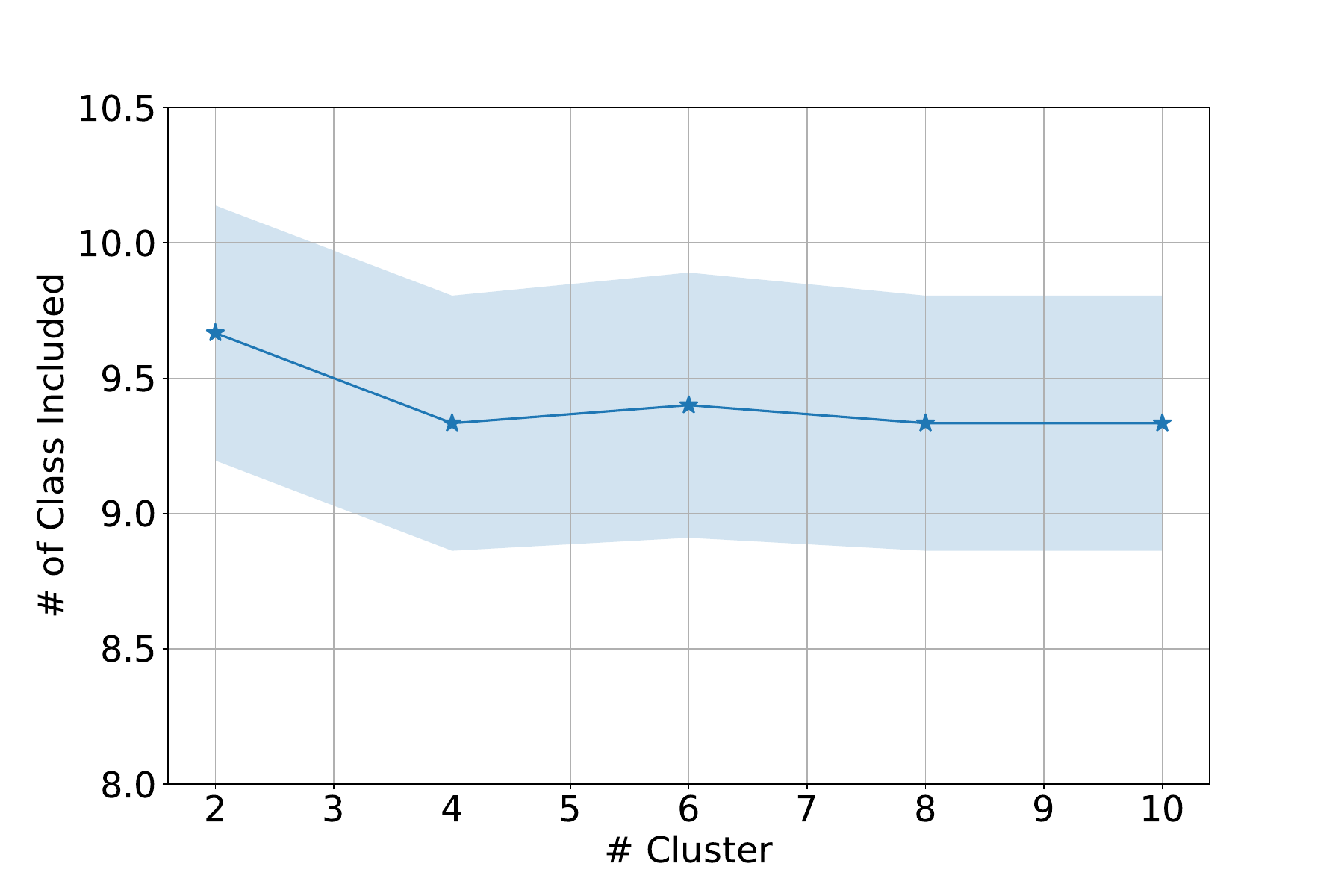}
    \caption{Class Coverage}
    \label{fig:class cover hyper}
  \end{subfigure}

  \caption{The impact of the number of clusters $C$ on the active learning strategy. Class coverage is robust to $C$, while larger $C$ means more accurate $y_{prior}$.}
  \label{fig:hyperC_c10}
\end{figure}

We investigated the impact of setting different clustering times, $C$, on the samples selected by our proposed active learning strategy. The experiment is implemented on CIFAR-10 with 10 labels. As shown in Fig.~\ref{fig:prior acc hyper} and Fig.~\ref{fig:class cover hyper}, we note that our strategy is robust to the number of clustering $C$. Across different values of $C$, our active learning strategy consistently selects samples that comprehensively cover all classes, even in scenarios with very few annotations. The choice of $C$ primarily influences the accuracy of the PPL. The larger $C$ is, the smaller the variance, and the accuracy of pseudo-labels can be slightly improved.



\subsubsection{Influence of Active Sampling Strategy}

We compare different active learning strategies on CIFAR-10, CIFAR-100, and STL-10, with all hyper-parameters consistent with those mentioned in Sec.~\ref{sec:setup}, except for the training iterations. In these experiments, all models are trained for 112$\times$2304 iterations. Many active learning strategies involve multiple rounds of iteration for selecting annotated samples, often resulting in high training costs, contradicting the fundamental goal of active semi-supervised learning, which aims to reduce the overall cost. Therefore, we choose active learning methods that allow for the selection of annotated samples in a single iteration as our baseline. Specially, random, K-medoids~\cite{sener2018active} and Coreset-greedy~\cite{sener2018active} are selected. As shown in tab.~\ref{table:al_acc}, our method consistently outperforms other active learning strategies. K-medoids outperforms random sampling in most cases, whereas Coreset often performs worse. This discrepancy arises because Coreset selects samples that are least similar to the features of the existing labeled samples, while K-medoids selects samples closest to the cluster centers. Consequently, Coreset is more likely to select outliers, which decreases the performance of semi-supervised training. This observation aligns with previous work~\cite{mittal2019parting}, which found that some active learning strategies can have adverse effects when the number of labeled samples is limited.

\begin{table}
\small
\begin{center}
\caption{Comparison of active sampling strategies. The accuracy reported is the average over 3 runs. The best results are shown in red and the second best results are shown in blue.}
\label{table:al_acc}
\begin{tabular}{l|ll|ll|ll}
\hline\noalign{}
Dataset & \multicolumn{2}{c|}{CIFAR-10} & \multicolumn{2}{c|}{CIFAR-100} & \multicolumn{2}{c}{STL-10} \\
Size of Labeled Set & 10 & 40 & 200 & 400 & 20 & 40 \\
\hline
Random & \color[rgb]{0,0,1}{58.17$\pm$13.01} & 82.81$\pm$8.10 & 38.84$\pm$3.09 & 58.42$\pm$2.37 & 51.14$\pm$6.79 & 63.33$\pm$5.97 \\
K-medoids & 47.26$\pm$6.62 & \color[rgb]{0,0,1}{87.42$\pm$5.42} & \color[rgb]{0,0,1}{43.39$\pm$4.00} & \color[rgb]{0,0,1}{59.10$\pm$0.36} & \color[rgb]{0,0,1}{51.03$\pm$4.89} & \color[rgb]{0,0,1}{68.29$\pm$2.93} \\
Coreset & 31.92$\pm$3.14 & 86.19$\pm$10.78 & 27.59$\pm$4.63 & 47.77$\pm$3.22 & 45.75$\pm$3.95 & 51.41$\pm$3.74 \\
Proposed & \color[rgb]{1,0,0}{84.43$\pm$5.19} & \color[rgb]{1,0,0}{94.25$\pm$0.43} & \color[rgb]{1,0,0}{51.22$\pm$3.06} & \color[rgb]{1,0,0}{61.07$\pm$0.12} & \color[rgb]{1,0,0}{61.39$\pm$5.43} & \color[rgb]{1,0,0}{70.26$\pm$3.60} \\
\hline
\end{tabular}
\end{center}
\end{table}

\subsubsection{Prior Pseudo-label Propagation}
For PPL generation, we compare our method with LLGC~\cite{zhou2003learning}, which is a typical baseline for label spreading, with the hyper-parameters of LLGC following~\cite{iscen2019label}. Additionally, we investigate the impact of selecting labeled samples using different active learning methods on the accuracy of pseudo-labels, as presented in Table~\ref{table: pseudo label ablation}. The results confirm that the choice of labeled sample selection strategy has a large impact on the accuracy of pseudo-labels. Samples selected by our active learning strategy result in more accurate pseudo-labels across different label propagation methods. Our pseudo-label propagation method outperforms LLGC when the number of labeled samples is close to the number of true classes, but shows slightly weaker performance than LLGC when more labels are available.

\begin{table}
\small
\begin{center}
\caption{Accuracy of prior pseudo-label comparison of different active strategies and label propagation methods. The accuracy reported is the average over 3 runs. The best results are shown in red and the second best results are shown in blue.}
\label{table: pseudo label ablation}
\begin{tabular}{ccccc}
\hline
\multicolumn{2}{c}{Dataset} & \multicolumn{3}{c}{CIFAR-10}  \\
\multicolumn{2}{c}{\# Labels} & 10 & 20 & 40  \\
Propagation & Sampling & \\
\hline\noalign{\smallskip}
LLGC & Random & 53.42$\pm$5.36 & 59.91$\pm$2.58 & 69.59$\pm$2.35  \\
LLGC & Coreset & 31.57$\pm$6.31 & 56.06$\pm$3.35 & 74.47$\pm$5.17 \\
LLGC & K-medoids & 45.53$\pm$2.96 & 62.28$\pm$5.39 & 71.51$\pm$1.36  \\
LLGC & USL & 53.90$\pm$0.56 & 63.47$\pm$3.19 & 70.82$\pm$1.76 \\
LLGC & Proposed & \color[rgb]{0,0,1}{62.94$\pm$3.47} & \color[rgb]{0,0,1}{71.61$\pm$1.64} & 
\color[rgb]{1,0,0}{75.50$\pm$1.40}  \\
Proposed & Random & 52.12$\pm$9.10 & 60.32$\pm$4.67 & 69.40$\pm$2.49 \\
Proposed & Coreset & 37.43$\pm$3.24 & 59.56$\pm$3.30 & 73.82$\pm$3.01  \\
Proposed & K-medoids & 44.90$\pm$1.88 & 62.12$\pm$6.97 & 69.62$\pm$0.83  \\
Proposed & USL & 54.20$\pm$0.43 & 57.62$\pm$5.10 & 67.61$\pm$1.86 \\
Proposed & Proposed & \color[rgb]{1,0,0}{71.41$\pm$4.10} & \color[rgb]{1,0,0}{72.63$\pm$0.96} & \color[rgb]{0,0,1}{74.91$\pm$1.66}  \\
\hline
\end{tabular}
\end{center}
\end{table}

Furthermore, we investigate the impact of the number of classes in clustering, $K$, in our label propagation. The experiments include three settings with different sizes of $K$, as summarized in Table~\ref{table: ablation of k}. Expected calibration error (ECE)~\cite{naeini2015obtaining} is employed to assess how well $y_{prior}$ is calibrated to the true accuracy, where smaller values indicate less miscalibration. The results confirm that using different $K$ in label propagation is a good compromise between accuracy and calibration.

\begin{table}
\small
\begin{center}
\caption{Ablation study of $K$ on CIFAR-10, accuracy of $y_{prior}$ and Expected calibration error (ECE) reported is the average over 5 runs.}
\label{table: ablation of k}
\begin{tabular}{llll|lll}
\hline
\quad & \multicolumn{3}{c|}{ECE} & \multicolumn{3}{c}{Accuracy of $y_{prior}$} \\
\# Labels & 10 & 20 & 40 & 10 & 20 & 40 \\
\hline
K=[10,10,10,10,10,10] & 0.278$\pm$0.039 & 0.166$\pm$0.038 & 0.112$\pm$0.021 & 70.50$\pm$4.13 & 70.66$\pm$4.05 & 70.81$\pm$2.13 \\
K=[10,20,30,40,50,60] & 0.237$\pm$0.041 & \color[rgb]{1,0,0}{0.087$\pm$0.014} & \color[rgb]{1,0,0}{0.067$\pm$0.013}&  \color[rgb]{1,0,0}{71.41$\pm$4.10} & \color[rgb]{1,0,0}{72.63$\pm$0.96} & \color[rgb]{1,0,0}{74.91$\pm$1.66} \\
K=[10,60,60,60,60,60] & \color[rgb]{1,0,0}{0.233$\pm$0.040} & 0.133$\pm$0.037 & 0.082$\pm$0.011 & 70.53$\pm$4.09 & 70.46$\pm$2.82 & 73.67$\pm$1.67 \\
\hline
\end{tabular}
\end{center}
\end{table}

\subsubsection{Influence of Self-Supervised Learning Tasks}

{We compared the impact of different self-supervised training methods on prior pseudo-label accuracy on CIFAR-10. Considering the availability of many pre-trained self-supervised models on large datasets in the deep learning community, we also investigated whether our method can generate high-quality prior pseudo-labels using these pre-trained models (i.e. model pre-trained on different datasets). As shown in Table~\ref{table: ssl_pplacc}, except for the image reconstruction-based pre-training method MAE~\cite{he2022masked}, our method (using clustering and label propagation) produces high-quality prior pseudo-labels across various contrastive self-supervised pre-training tasks.

\begin{table}[!htb]
\small
\begin{center}
\caption{The impact of different self-supervised learning tasks on prior pseudo-label accuracy on CIFAR-10. Results are averaged over 3 runs. The best results are shown in red and the second best results are shown in blue.}
\label{table: ssl_pplacc}
\begin{tabular}{lllllll}
\hline
Self-sup. Task & Pre-trained Data & Model & 10 Labels & 20 Labels & 40 Labels \\
\hline\noalign{\smallskip}
Simsiam~\cite{chen2021exploring} & CIFAR-10 & WRN-28-2 & 71.41$\pm$4.10 & 72.63$\pm$0.96 & 74.91$\pm$1.66 \\
SimCLR~\cite{chen2020simple} & CIFAR-10 & WRN-28-2 & \color[rgb]{0,0,1}{72.63$\pm$3.40} & \color[rgb]{0,0,1}{75.95$\pm$0.13} & \color[rgb]{0,0,1}{78.45$\pm$0.28}  \\
BYOL~\cite{grill2020bootstrap} & ImageNet & ResNet-50 & 49.27$\pm$3.89 & 64.13$\pm$0.55 & 67.80$\pm$1.61  \\
MAE~\cite{he2022masked} & ImageNet & ViT-B/16 & 29.18$\pm$0.54 & 32.61$\pm$0.24 & 35.98$\pm$0.23  \\
CLIP~\cite{radford2021learning} & WebImageText & ViT-B/32 &  \color[rgb]{1,0,0}{75.94$\pm$0.07} &  \color[rgb]{1,0,0}{80.77$\pm$2.79} &  \color[rgb]{1,0,0}{84.19$\pm$0.32}  \\
\hline
\end{tabular}
\end{center}
\end{table}

\subsubsection{Clustering Results on Self-Supervised Features}
\label{sec:cluster_self}

We analyzed clustering results in self-supervised feature space, as shown in Fig.~\ref{fig:cluster_self_c10} and Fig.~\ref{fig:cluster_self_c100}. The self-supervised features were clustered five times, with the number of clusters equal to the actual number of classes in the datasets. The samples were divided into 10 bins based on the distance between the sample features and the cluster centers. The proportion of sample labels in each bin matching the dominant labels in the clusters was then calculated. The result shows that samples near the center of the cluster are more likely to share the same label as the dominant label in the cluster.



\begin{figure}[!htbp]
  \centering
  
  \begin{subfigure}{0.49\textwidth}
    \includegraphics[width=\linewidth]{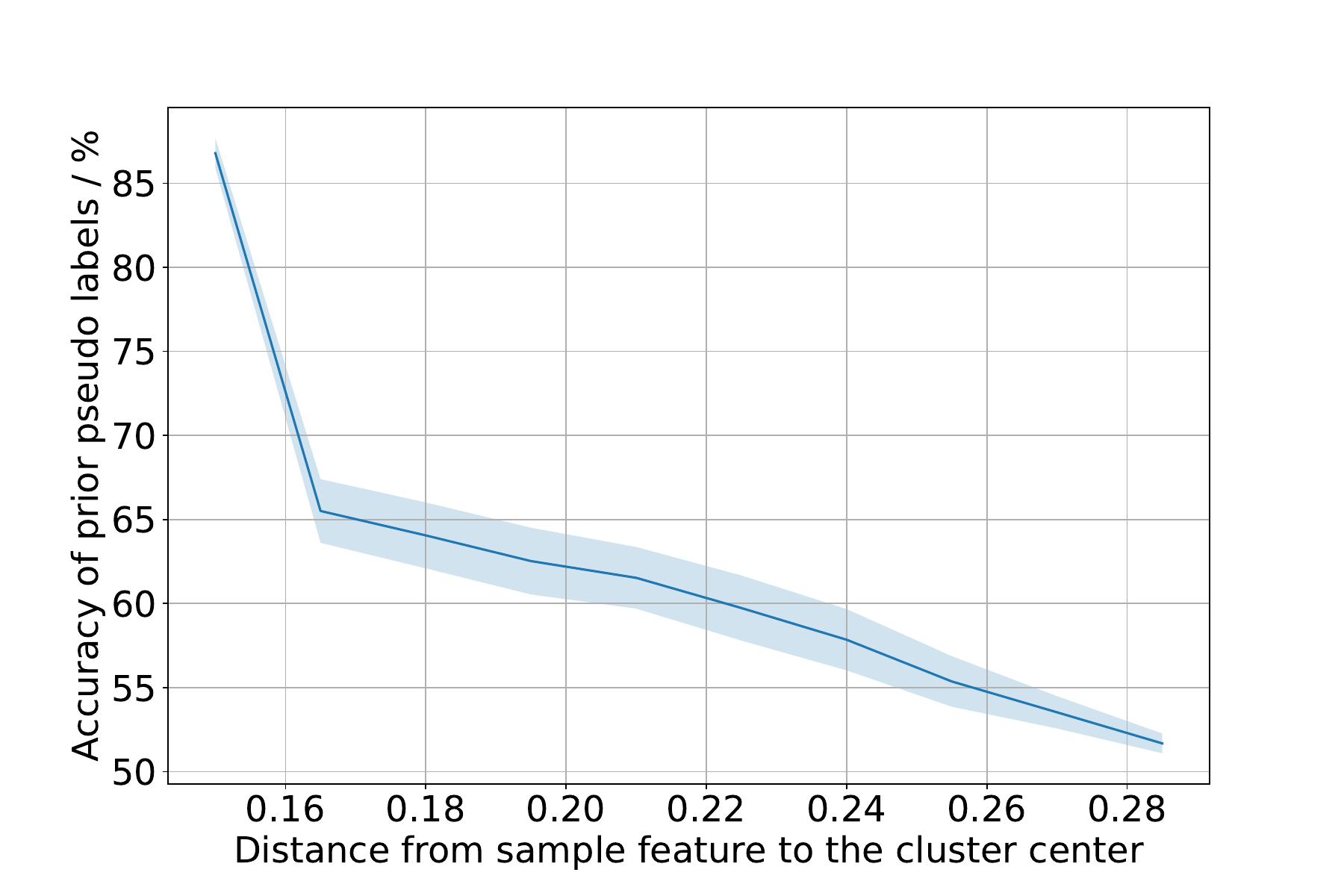}
    \caption{CIFAR-10}
    \label{fig:cluster_self_c10}
  \end{subfigure}
  \begin{subfigure}{0.49\textwidth}
    \includegraphics[width=\linewidth]{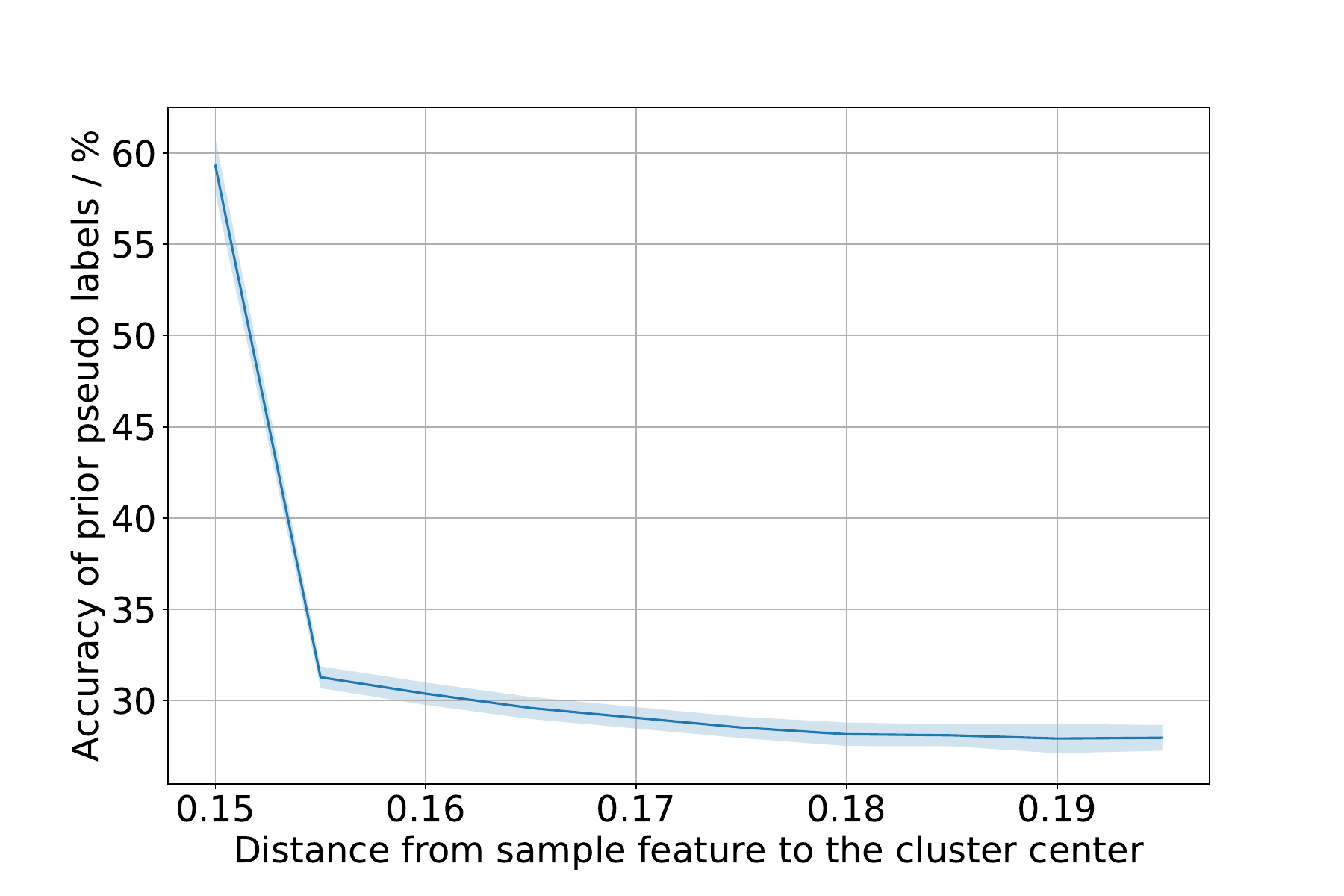}
    \caption{CIFAR-100}
    \label{fig:cluster_self_c100}
  \end{subfigure}

  \caption{The relationship between distance and dominant labels in self-supervised feature space, where the model is self-supervised training by Simsiam~\cite{chen2021exploring}.}
  \label{fig:cluster_self}
\end{figure}

\section{Conclusions and Limitations}
In this paper, we observed that in scenarios with limited labeled samples, semi-supervised training often quickly compromises the well-established feature representations obtained through self-supervised training. Weight initialization, in such cases, fails to effectively transfer valuable information from self-supervised pre-training to the semi-supervised model, resulting in a reduction of the benefits gained from combining self-supervised and semi-supervised training. Motivated by this observation, we propose using PPL as an intermediate step to assist semi-supervised training in acquiring valuable information from self-supervised pre-training. Additionally, to more fully exploit self-supervised pre-training information, we introduce a novel active learning strategy to enhance the accuracy of PPL. By combining weight initialization, PPL, and the active learning strategy, our framework provides a more effective starting point for semi-supervised training. Experiments on multiple image classification datasets demonstrate the effectiveness of our approach in enhancing semi-supervised learning performance with limited labeled data. Furthermore, our method readily integrates with existing semi-supervised learning approaches that include pseudo-labeling techniques.

Our method primarily relies on prior pseudo-labels generated through clustering on pre-trained features to enhance semi-supervised training performance. This approach faces limitations in scenarios where clustering cannot effectively distinguish between different classes, such as in multi-label learning where a single image contains multiple objects. Additionally, the fixed pseudo-label switching time $T$ may hinder performance in nonstationary environments, such as stream learning.

\section{Acknowledgements}

The authors acknowledge the University of Sydney’s high performance computing cluster, Artemis, for providing the computing resources. This work was supported by the Australian Research Council [Grant LE200100049].



\bibliographystyle{elsarticle-num} 
\bibliography{bib}





\subsection*{  } 

\setlength\intextsep{-12pt} 
\begin{wrapfigure}{l}{0.13\textwidth}
    \centering
    \includegraphics[width=0.15\textwidth]{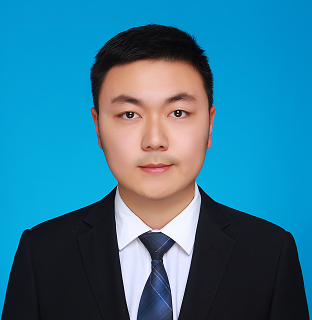}
\end{wrapfigure}
\noindent \textbf{Ziting Wen} received a BEng in Electronic Science and Technology in Southeast University in 2017 and a MSc in Information and Communication Engineering in Shanghai Jiao Tong University in 2020. He is currently pursuing the PhD in University of Sydney’s Australian Centre for Field Robotics. His research interests include label-efficient learning and computer vision. 

\subsection*{  } 
\setlength\intextsep{-3pt} 
\begin{wrapfigure}{l}{0.13\textwidth}
    \centering
    \includegraphics[width=0.15\textwidth]{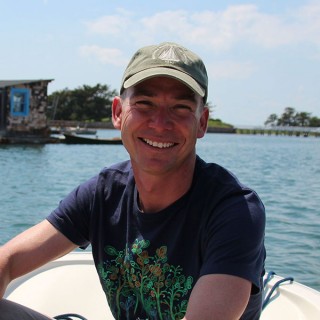}
\end{wrapfigure}
\noindent \textbf{Oscar Pizarro} received a BSc in electronic engineering from the Universidad de Concepcion in 1997, a dual MSc in ocean engineering and electrical engineering and computer science and PhD in oceanographic engineering from the MIT-WHOI Joint Program,in 2003 and 2004, respectively. He joined the University of Sydney’s Australian Centre for Field Robotics in 2005, where he is Principle Research Fellow. His research interests include scalable approaches to seafloor imaging and habitat characterisation.

\subsection*{  } 
\setlength\intextsep{-3pt} 
\begin{wrapfigure}{l}{0.13\textwidth}
    \centering
    \includegraphics[width=0.15\textwidth]{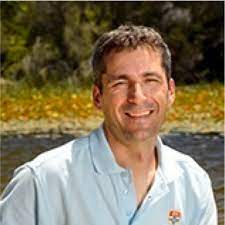}
\end{wrapfigure}
\noindent \textbf{Stefan Williams} received a BSc in systems engineering design from the University of Waterloo in 1997 and a PhD in field robotics from the University of Sydney in 2002. He is Professor of Marine Robotics at the University of Sydney’s Australian Centre for Field Robotics, and heads Australia’s Integrated Marine Observing System AUV Facility. His research interests include simultaneous localization and mapping in underwater environments, autonomous navigation and data interpretation.

\end{document}